\documentclass[sigconf]{acmart}
\AtBeginDocument{%
  }

\usepackage{graphicx}
\usepackage{booktabs}
\usepackage{algorithm}
\usepackage{algorithmic}

\usepackage{amssymb}
\usepackage{amsfonts}
\usepackage{comment}
\usepackage{multirow}
\usepackage{xcolor} % 提供颜色支持
\usepackage{booktabs} % 用于 \toprule, \midrule, \bottomrule
\usepackage{wrapfig} % 文本环绕式表格
\usepackage{rotating} % 表格旋转
\usepackage{cleveref}
\usepackage{caption}
\usepackage{array}
\usepackage{hyperref}

\copyrightyear{2026}
\acmYear{2026}
\setcopyright{cc}
\setcctype{by}
\acmConference[MM '26] {Proceedings of the 34th ACM International Conference on Multimedia}{November 10--14, 2026}{Rio de Janeiro, Brazil.}
\acmDOI{10.1145/3767308.3835322}

\acmBooktitle{Proceedings of the 34th ACM International Conference on Multimedia (MM '26), November 10--14, 2026, Rio de Janeiro, Brazil}
\acmISBN{979-8-4007-2213-4/2026/11}

\makeatletter
\newcommand{\correspondingauthornote}[1]{%
  \g@addto@macro\addresses{\@@authornotemark{2}}%
  \g@addto@macro\@authornotes{\footnotetext[2]{#1}}%
}
\makeatother

\begin{document}


\title{Open-Linguistic Concept Unified Learning for Cross-Site Interpretable Dermatology Image Diagnosis}

\author{Chengyu Wu}
\orcid{https://orcid.org/0009-0009-0450-8649}
\affiliation{
\institution{Zhejiang University}
        \city{Hangzhou}
	\country{China}
    }
\affiliation{%
	\institution{Westlake University}
        \city{Hangzhou}
	\country{China}
    }
\email{wuchengyu@westlake.edu.cn}

\author{Junpeng Tan}
\orcid{https://orcid.org/0000-0001-9546-4331}
\affiliation{%
	\institution{Westlake University}
        \city{Hangzhou}
	\country{China}}
\email{tanjunpeng@westlake.edu.cn}

\author{Wanxiang Luo}
\orcid{https://orcid.org/0009-0000-5174-8535}
\affiliation{%
  \institution{Harbin Institute of Technology}
  \city{Shenzhen}
  \country{China}
}
\email{25B965023@stu.hit.edu.cn}

\author{Yaqi Wang}
\orcid{https://orcid.org/0000-0002-4627-3392}
\affiliation{%
 \institution{Hangzhou Dianzi University}
 \city{Hangzhou}
 \country{China}}
\email{wangyaqi@hdu.edu.cn}

\author{Yandong Wen}
\orcid{https://orcid.org/0000-0001-6330-7438}
\affiliation{%
  \institution{Westlake University}
        \city{Hangzhou}
	\country{China}}
\email{wenyandong@westlake.edu.cn}

\author{Yefeng Zheng}
\orcid{https://orcid.org/0000-0003-2195-2847}
\correspondingauthornote{Corresponding author.}
\affiliation{%
	\institution{Westlake University}
        \city{Hangzhou}
	\country{China}
}
\email{zhengyefeng@westlake.edu.cn}

\renewcommand{\shortauthors}{Chengyu Wu et al.}


\begin{abstract}
Human-interpretable computer-aided diagnosis is crucial for clinical decision making. Concept-based models excel by providing transparent reasoning and enabling post-hoc, clinician-in-the-loop interventions. However, their rigid dataset-specific adaptation inherently restricts cross-site generalization. Applying them across diverse modalities, such as dermoscopic and clinical photographs, is challenging due to heterogeneous concept taxonomies varying in availability, granularity, and semantics across cohorts. Consequently, adapting Foundation Vision-Language Models (FVLMs) demands costly label engineering and repeated post-training. Existing intervention mechanisms remain rigidly tied to predefined concepts, lacking adaptability and hindering scalable dermatology CAD deployment. To address these bottlenecks, we propose UniCon, an open-linguistic unified concept learning framework for multimodal interpretable vision-language diagnosis. UniCon resolves these challenges through three contributions: (1) A shared semantic representation space via a unified concept prototype codebook, seamlessly coordinating heterogeneous concept systems across modalities without dataset-specific retraining. (2) Open-linguistic based multi-faceted semantic specifications to overcome sparse textual label limitations, improving boundary sensitivity in uncertain clinical contexts. (3) A robust, cross-site adjustable intervention interface powered by reliability-gated bottleneck aggregation, enabling consistent reasoning and transferable clinician corrections. Extensive experiments demonstrate that beyond securing top-tier diagnostic accuracy, UniCon successfully bridges disparate clinical taxonomies, unlocking unprecedented cross-site intervention capabilities. Code is available at \url{https://github.com/wuchengyu123/UniCon}.
\end{abstract}

% Abstract 的逻辑： (1) 码本解决统一空间；(2) 开放语义解决文本稀疏；(3) 门控聚合解决干预接口。







\maketitle

\section{Introduction}
\label{sec:intro}

Recent advances in deep learning have significantly improved Computer-Aided Diagnosis (CAD) systems, evolving from manual feature engineering to end-to-end deep neural network architectures~\cite{han2022survey}. These systems now match or exceed clinical expert performance in tasks such as diabetic retinopathy detection~\cite{gulshan2016development}, skin cancer classification~\cite{esteva2017dermatologist}, and lung nodule screening~\cite{ardila2019end}, positioning CAD as a viable clinical decision support tool. However, most current methods prioritize prediction accuracy while overlooking critical clinical requirements: model transparency~\cite{chen2022interpretable}, reasoning credibility~\cite{mudgal2020ethical}, error traceability~\cite{wu2024semi,Gong2022}, and human-in-the-loop intervention capabilities~\cite{matta2024systematic,wu2025samvsr}.

\begin{table}[htbp]
    \centering
    % 调整行高
    \renewcommand{\arraystretch}{1.5} 
    % 缩小表格整体字体，避免在狭窄列宽下文字换行过于频繁
    \small 
    \caption{Concepts of various dermatological imaging}
    % 调整各列宽，总和约 8.2cm，刚好适配常见的单栏宽度 (\linewidth)
    \begin{tabular}{
        >{\centering\arraybackslash}m{1.8cm} 
        >{\centering\arraybackslash}m{2.0cm}
        >{\centering\arraybackslash}m{3.5cm}
    }
        \toprule
        \textbf{Images} & \textbf{Image Type} & \textbf{Concepts} \\
        \midrule
        
        % 第一行：图 (a) 及其对应数据
        \vspace{0.05cm} % 稍微缩小顶部微调距离
        \includegraphics[width=1.0cm]{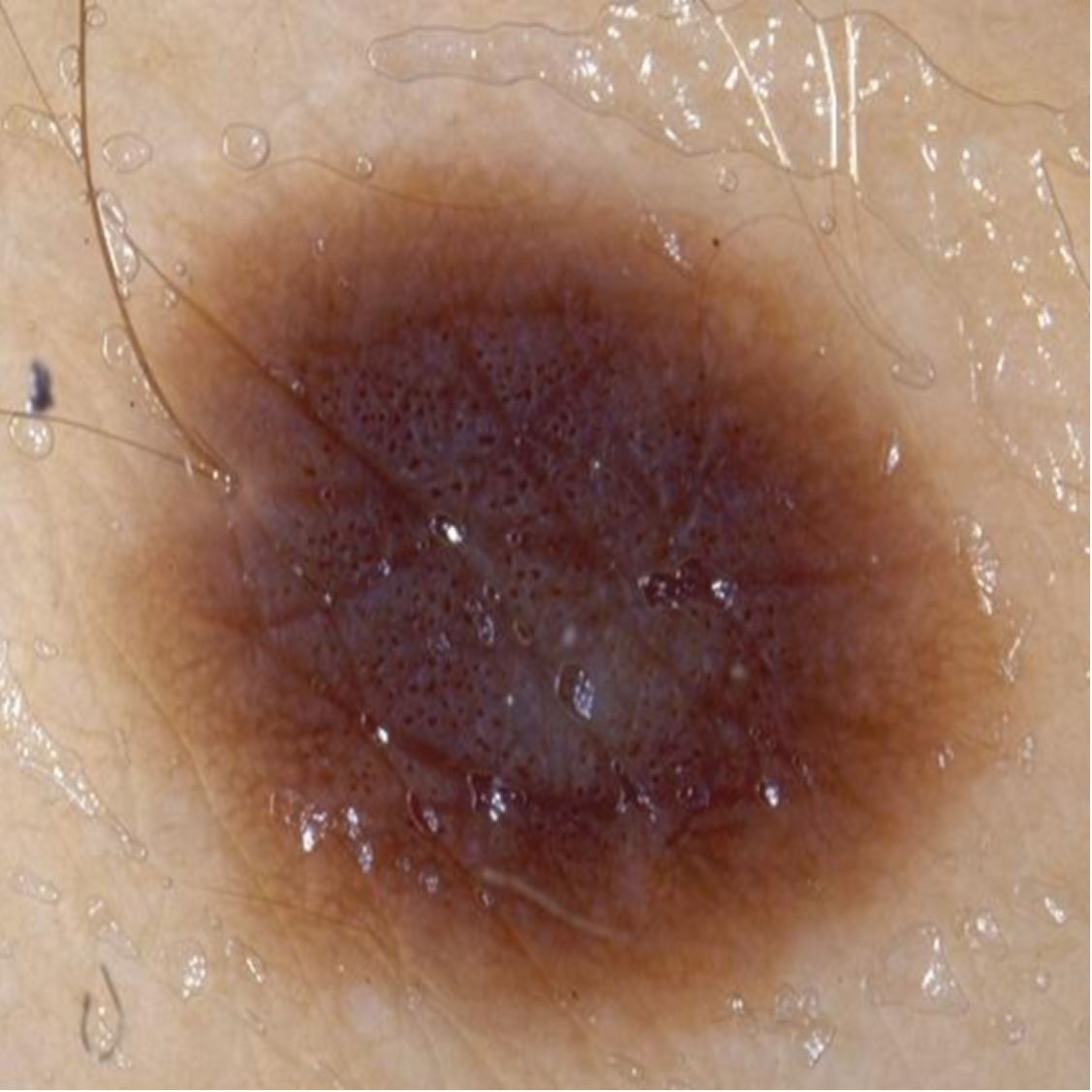}  & 
        Demoscopic images & 
        Pigment network, Streaks, Blue-whitish veil... \\
        
        \midrule
        
        % 第二行：图 (b) 及其对应数据
        \vspace{0.05cm}
        \includegraphics[width=1.0cm]{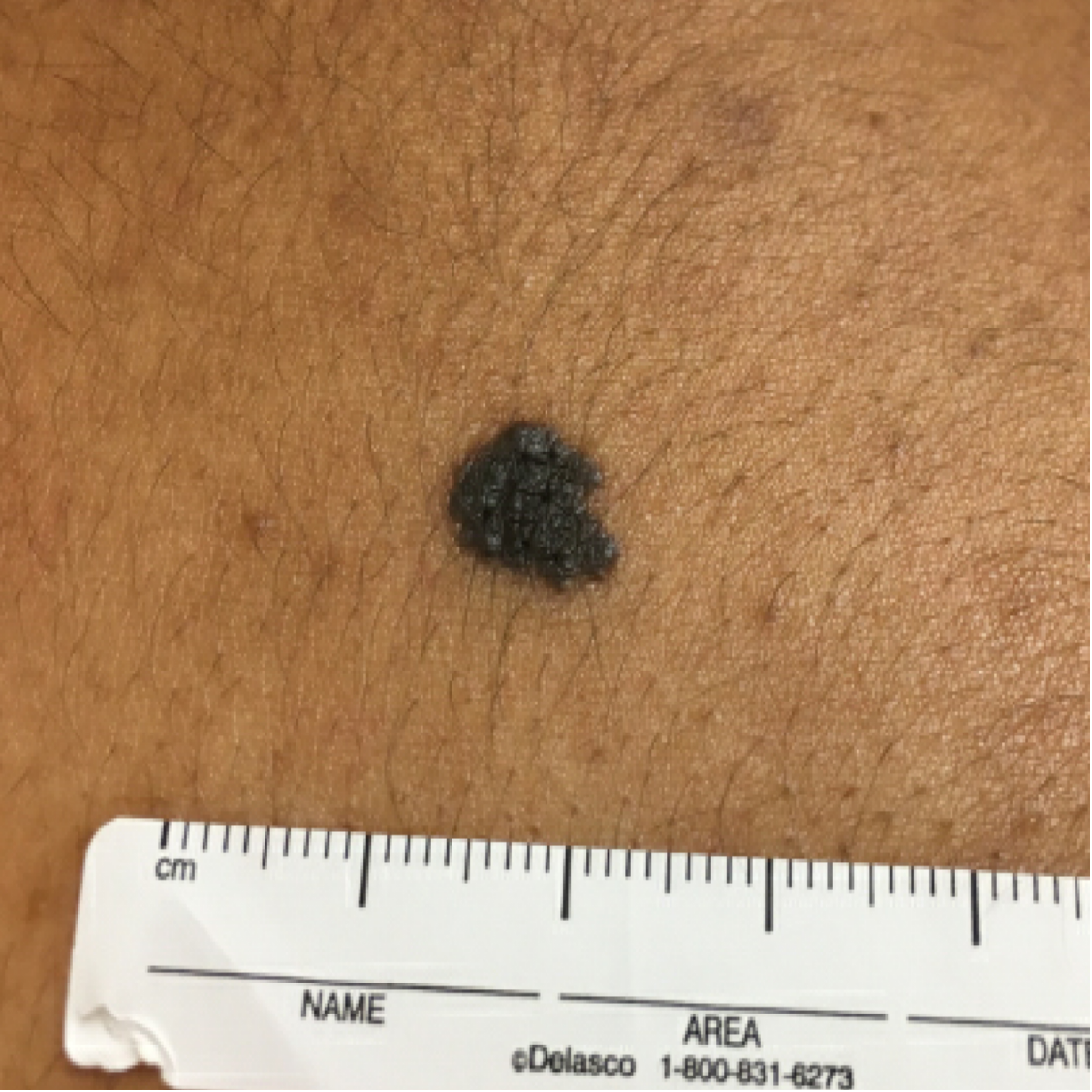} & 
        Clinical images & 
        Plaque, Patch, Nodule, Crust, Yellow... \\
        
        \bottomrule
    \end{tabular}
    \label{intro_skin_images}
\end{table}

Dermatological diagnosis presents unique challenges distinct from diabetic retinopathy or lung nodule detection, where anatomical structures provide stable spatial references~\cite{gulshan2016development,ardila2019end}. Skin lesions appear anywhere on the body with highly variable morphology, exhibiting subtle inter-class differences yet substantial intra-class variability due to various reasons~\cite{esteva2017dermatologist}, resulting in a feature space with high inter-class similarity and intra-class diversity. Moreover, while clinical dermatology relies on structured attribute-based rules~\cite{wang2025integrating}, end-to-end models typically output opaque class probabilities without explicit modeling of visual evidence or intermediate reasoning. This disconnect compromises interpretability, auditability, and human-in-the-loop intervention, necessitating evidence-based, clinically verifiable prediction systems.

\begin{figure}[tb]
  \centering \includegraphics[width=0.9\linewidth]{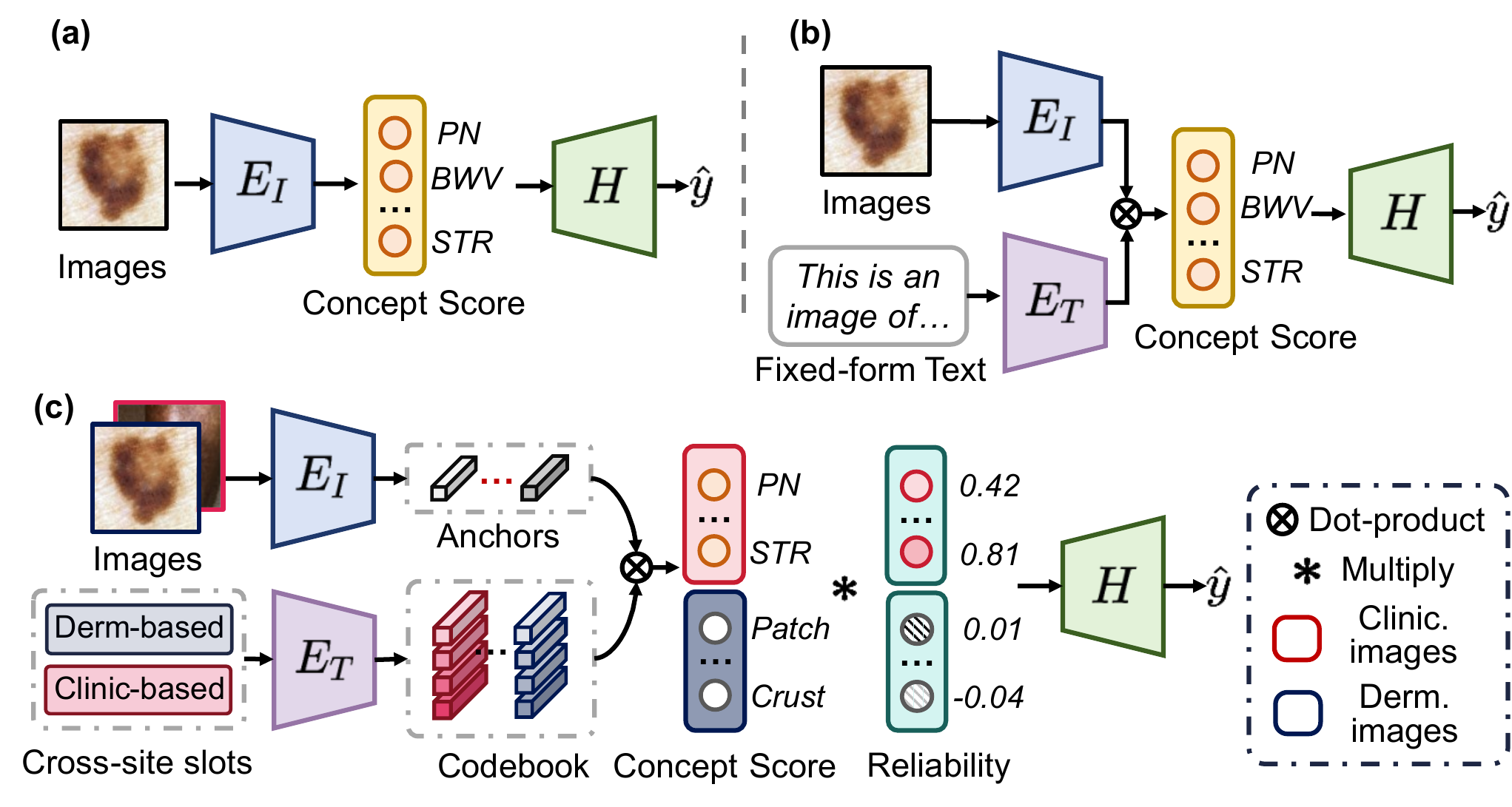}
  \caption{Overview of existing approaches versus ours.
(a) CBM~\cite{koh2020concept}: fixed handcrafted concepts.
(b) CBM-CLIP~\cite{patricio2024towards}: fixed text templates.
(c) Ours: multi-modal concepts with reliability-aware diagnosis.
  }
  \label{fig:intro_method}
\end{figure}

Concept Bottleneck Models (CBMs)~\cite{koh2020concept} advance interpretable diagnosis by encoding images into human-understandable clinical concepts before final prediction~\cite{Fonseca2024Concept}, enabling transparent post-hoc interventions. However, cross-site deployment is severely hindered by concept heterogeneity and varying acquisition protocols~\cite{wen2022characteristics}. As shown in Table~\ref{intro_skin_images}, professional dermoscopes use illuminated magnification to eliminate reflection and analyze sub-surface micro-structural concepts (e.g., Pigment Network)~\cite{kawahara2018seven}, whereas standard clinical photographs rely on macroscopic concepts (e.g., Plaque) to describe visible surface morphology~\cite{daneshjou2022skincon}. While Foundation Vision-Language Models (FVLMs) offer a promising language-supervised to generalizing concept learning~\cite{moor2023foundation,oikarinen2023label} for scalable diagnosis~\cite{yang2023language}, stronger pretraining alone cannot inherently resolve this fundamental structural mismatch in semantic space~\cite{oakden2020hidden}. Because attribute taxonomies and hierarchies naturally differ across sites, adapting FVLMs still necessitates costly dataset-specific engineering and repeated adaptation~\cite{willemink2020preparing,sun2026quota}, which impedes scalable clinical deployment~\cite{finlayson2021clinician,wiggins2022opportunities}. Consequently, existing approaches remain constrained by the lack of a unified, transferable semantic framework, limiting their cross-site generalization and clinical deployment.

\begin{figure}[!tb]
  \centering \includegraphics[width=0.9\linewidth]{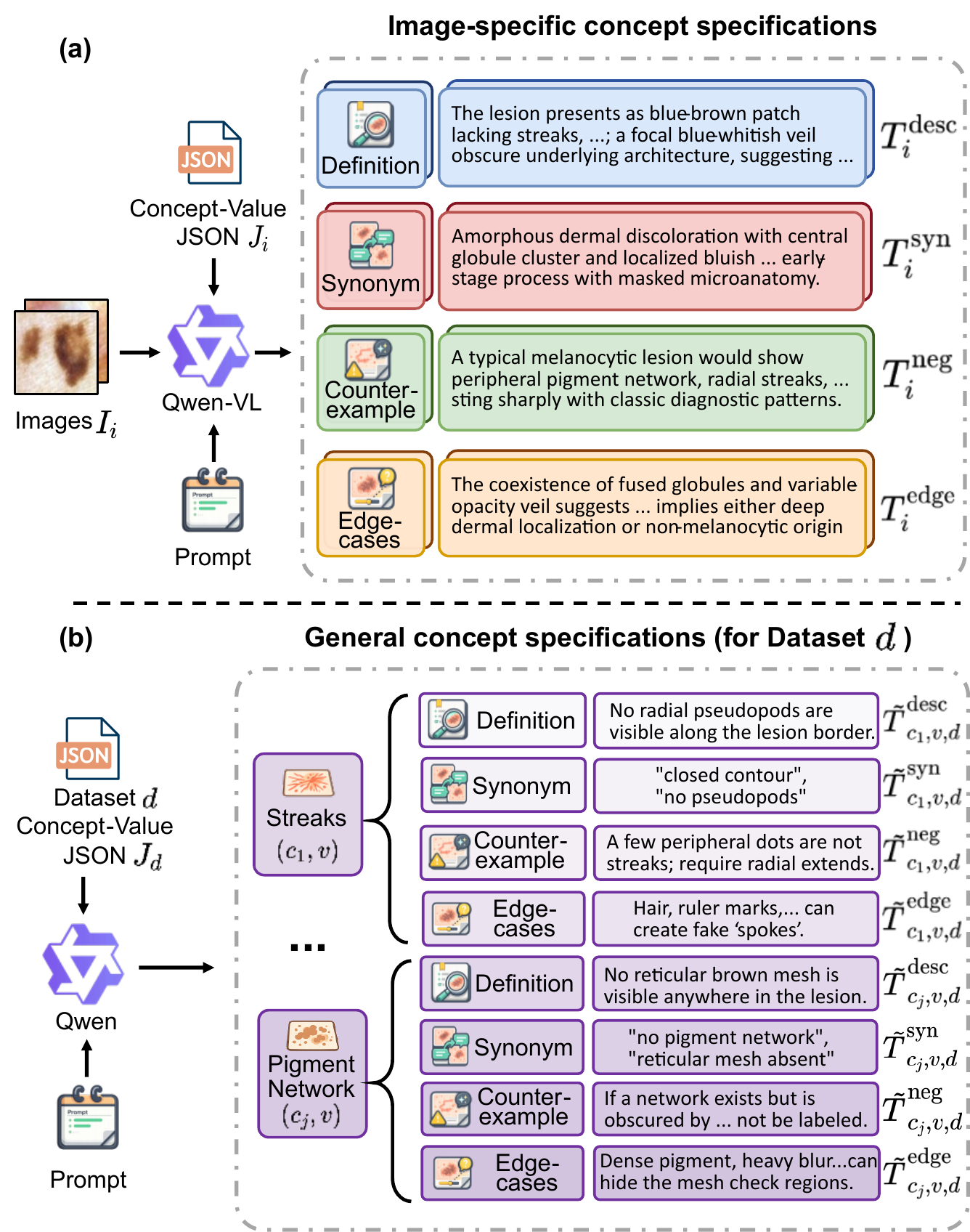}
  \caption{Illustration of the proposed Multi-Faceted Semantic Specifications (MSS). (a) Instance-level textual descriptions ($T_i^x$) prompted from specific images; (b) Concept-level textual descriptions ($\tilde{T}_{c,v,d}^x$) constructed for unified concept representation.
  }
  % \vspace{-0.3cm}
  \label{fig:example}
\end{figure}

To address these critical bottlenecks, we propose UniCon, an open-linguistic unified concept learning framework for multimodal interpretable vision-language diagnosis. As shown in Fig.~\ref{fig:intro_method}, unlike prior methods that rely on isolated, dataset-specific concept sets~\cite{koh2020concept}, UniCon establishes a shared semantic space across heterogeneous modalities (dermoscopic and clinical images). It introduces a standardized training-inference-intervention paradigm that enables robust, cross-site diagnostic reasoning. Crucially, this approach achieves seamless generalization across heterogeneous clinical cohorts and imaging modalities, remaining entirely agnostic to domain origins and obviating dataset-specific fine-tuning. The core contributions of this work are summarized as follows:

% 逻辑链：（1) 码本解决统一空间；(2) 开放语义解决文本稀疏；(3) 门控聚合解决干预接口 

(1) To overcome cross-site concept heterogeneity, we construct a unified concept prototype codebook that serves as a shared semantic representation space. By mapping disparate, site-specific concept taxonomies to this codebook, UniCon seamlessly coordinates heterogeneous concept systems across modalities, ensuring consistent model training and reasoning without dataset-specific adaptation or label engineering.

(2) To address the limitations of sparse textual semantics and ambiguous concept boundaries, we introduce multi-faceted, open-linguistic semantic specifications. This approach fundamentally enriches the linguistic expression of visual evidence in uncertain contexts, significantly improving boundary sensitivity and global semantic alignment between multimodal image-concept pairs.

(3) To break the rigidity of predefined, site-specific interventions, we propose a reliability-gated bottleneck aggregation strategy. This gating mechanism provides a robust and transferable interface for clinician-guided interventions. It automatically assesses concept reliability, enabling consistent, adjustable, and transparent multi-modal reasoning across diverse clinical populations.

\section{Related Works}
\subsection{Concept Bottleneck Models}
Concept Bottleneck Models (CBMs) enhance interpretability by introducing an intermediate concept layer~\cite{yuksekgonul2022post}. However, traditional CBMs require dense annotations and complete retraining for vocabulary expansion~\cite{patricio2025two}, hindering dynamic clinical deployment. In dermatology, static concept sets are particularly vulnerable to concept heterogeneity—varying terminology and granularity across institutions~\cite{xu2025concept,pang2024integrating}. This degrades performance under shifting acquisition protocols, such as transitioning between dermoscopic and clinical images. Consequently, test-time interventions learned for one site rarely transfer to others~\cite{shin2023closer,shen2025adaptive}. Compounded by pretraining semantic misalignments~\cite{gao2024evidential,bie2024mica}, current CBMs fundamentally lack a unified representation accommodating cross-site variability and clinician-in-the-loop controllability.

\subsection{Vision-Language Models}

Vision-Language Models (VLMs) align visual features with clinical language for versatile cross-modal applications like zero-shot classification~\cite{radford2021learning, Gong2026, li2023blip, liu2023visual}. In dermatology, VLMs provide robust, clinically relevant explanations~\cite{chen2025evaluating,du2024ipo,Gong2024}. Despite these advantages, adapting VLMs introduces site-specific overhead; unique institutional vocabularies necessitate costly prompt engineering~\cite{du2024ipo}, impeding multi-center scalability~\cite{sun2023metamodulation}. Critically, VLMs lack a unified, interpretable control interface. Their opaque reasoning prevents clinicians from correcting errors via intermediate concepts~\cite{dutta2025vision}, exacerbating cross-site misinterpretations. Furthermore, VLMs' tendency to hallucinate and over-rely on prior knowledge~\cite{wolf2025your, dutta2025vision} underscores the pressing need for reliable, controllable human-in-the-loop mechanisms.

Unlike CBMs restricted by static vocabularies and VLMs constrained by opaque reasoning, our proposed UniCon harmonizes heterogeneous concepts into a shared semantic space, providing a transferable intervention interface for consistent corrections across diverse sites and imaging modalities.

\section{Methods}
Given multi-cohort skin lesion datasets $\{\mathcal{D}_d\}_{d=1}^{D}$, each sample comprises an image $I_{d_i}$, cohort-specific concepts $\{c^{d_i}_1,\ldots,c^{d_i}_j\}$, and diagnosis label $y$. A unified concept set $\mathcal{C}=\{c_1,\dots,c_{|\mathcal{C}|}\}$ establishes a shared semantic space. Instead of binary presence, each concept $c\in\mathcal{C}$ acts as a clinical attribute defined by a discrete state space $\mathcal{V}_c$ (e.g., for $c = \text{``pigment network''}$, $\mathcal{V}_c = \{\text{typical}, \text{atypical}, \text{absent}\}$).

\begin{figure*}[tb]
  \centering
\includegraphics[width=0.9\textwidth]{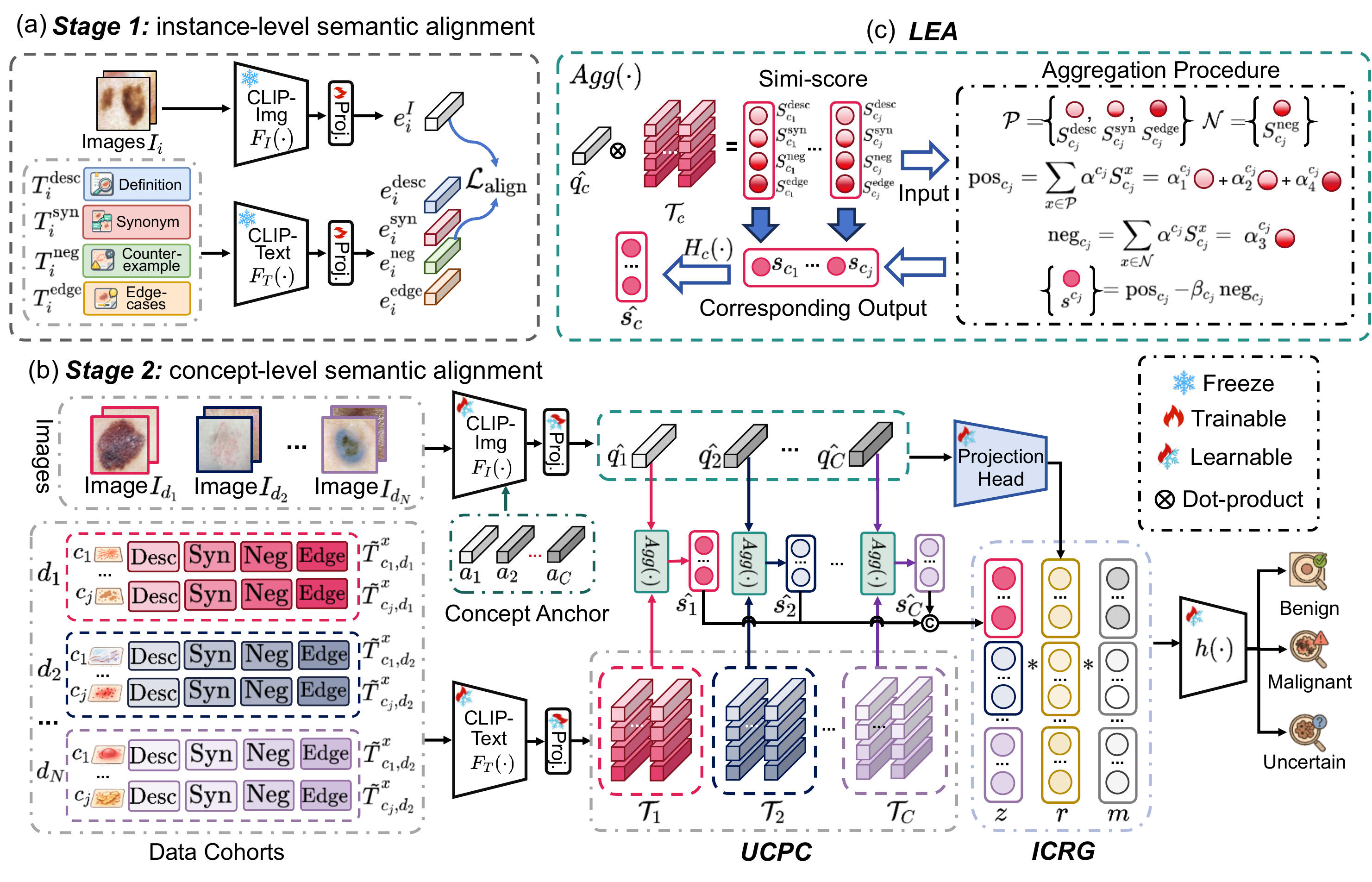}
\caption{Overview of UniCon. (a) Stage 1 aligns images and MSS in a shared space using the multi-margin objective $\mathcal{L}_{\text{align}}$. (b) Stage 2 uses cross-cohort texts $\tilde{T}_{c,d}^{x}$ to construct the UCPC. Visual queries $\hat{q}_c$ probe the codebook, while ICRG filters the concept bottleneck using reliability $r$ and applicability $m$ before $h(\cdot)$. (c) LEA aggregates positive $\mathcal{P}$ and negative $\mathcal{N}$ evidence from $\mathcal{T}_c$ into calibrated scores $\hat{s}_c$. All modules are jointly optimized in three steps.}
  \label{fig:main}
\end{figure*}

UniCon comprises two interconnected stages. \textbf{Stage 1} (Fig.~\ref{fig:main}(a)) aligns each image with four-slot Multi-Faceted Semantic Specifications (MSS), producing language-robust and boundary-sensitive instance representations. \textbf{Stage 2} (Fig.~\ref{fig:main}(c)) reuses the optimized text encoder and projector to aggregate cross-cohort concept-level MSS into a Unified Concept Prototype Codebook (UCPC), providing dataset-agnostic representations for every concept state. Learnable Evidence Aggregation (LEA) (Fig.~\ref{fig:main}(b)) and Image-Conditioned Reliability Gating (ICRG) (Fig.~\ref{fig:main}(c)) then enable interpretable semantic fusion by dynamically evaluating and filtering concept applicability, yielding a transferable bottleneck for cross-site diagnosis and clinician-in-the-loop intervention.

\subsection{Instance-Level Multi-Faceted Semantic Alignment}

In Stage 1, we employ contrastive optimization to construct a language-robust and boundary-sensitive semantic space. We achieve a global collaborative representation between the image and text encoders by utilizing the Multi-Faceted Semantic Specifications (MSS) generated for each specific image.

\noindent\textbf{Multi-Faceted Semantic Specifications (MSS).}
Instead of relying on single-template prompts, we utilize Qwen3-VL~\cite{bai2025qwen3} to generate four complementary textual slots for each image $I_i$ based on its static concept annotations $\{c_{i,1},\ldots,c_{i,j}\}$:
\begin{itemize}
    \item \textbf{Description (${T}_i^{\text{desc}}$):} A comprehensive and canonical definition of the target clinical concept, detailing its primary morphological features and standard diagnostic criteria.
    \item \textbf{Synonyms (${T}_i^{\text{syn}}$):} Alternative clinical terminologies to accommodate linguistic variations prevalent across heterogeneous medical cohorts.
    \item \textbf{Edge-cases (${T}_i^{\text{edge}}$):} Characterizations of atypical clinical presentations or ambiguous boundaries to enhance robustness in highly uncertain regions.
    \item \textbf{Counter-examples (${T}_i^{\text{neg}}$):} Explicit descriptions of visually analogous but pathologically distinct clinical signs, serving as critical negative constraints.
\end{itemize}

\noindent\textbf{Global Image-Text Contrastive Optimization.}
To achieve holistic instance-level alignment, we extract and $L_2$-normalize the visual embedding $e_{I_i}=\mathrm{Norm}(P_I(F_{\text{I}}(I_i))) \in \mathbb{R}^{d}$ using a frozen BiomedCLIP~\cite{zhang2023biomedclip} image encoder $F_{\text{I}}$ and a learnable projector $P_I$. Simultaneously, the four textual slots are encoded into instance-level semantic anchors $e_i^x = \mathrm{Norm}(P_T(F_{\text{T}}(T_i^x))) \in \mathbb{R}^{d}$ for $x \in \{\text{desc}, \text{syn}, \text{edge}, \text{neg}\}$, via a frozen text encoder $F_{\text{T}}$ and projector $P_T$. We optimize a structured multi-margin objective comprising three components:

\noindent(1) \textit{Intra-Textual Anchoring Constraints:} 
To ensure positive anchors $\mathcal{P}_i = \{e_i^{\text{desc}}, e_i^{\text{syn}}, e_i^{\text{edge}}\}$ form a compact semantic neighborhood, we apply a pair of margin constraints:
\begin{equation}
\mathcal{L}_{\text{intra}} = \sum_{u,w \in \mathcal{P}_i} \left[ \max(0,\|u-w\|_2-\tau_{\text{close}}) + \max(0,\tau_{\text{spread}}-\|u-w\|_2) \right].
\end{equation}

\noindent(2) \textit{Semantic-to-Visual Anchoring:} 
Let $\mu_i = (e_i^{\text{desc}} + e_i^{\text{syn}} + e_i^{\text{edge}})/\| e_i^{\text{desc}} + e_i^{\text{syn}} + e_i^{\text{edge}} \|$ be the semantic centroid. We encourage $e_{I_i}$ to fall into the region spanned by $\mathcal{P}_i$, while keeping it strictly bounded away from the counter-example $e_i^{\text{neg}}$:
\begin{equation}
\mathcal{L}_{\text{cross}} = \max(0, \|e_{I_i} - \mu_i\|_2 - \tau_{\text{in}}) + \max(0, \tau_{\text{away}} - \|e_{I_i} - e_i^{\text{neg}}\|_2).
\end{equation}

\noindent(3) \textit{Inter-Sample Class Separation:} 
To maintain inter-class discriminability, representations from different classes must be pushed apart. For any negative sample $I_j$ ($y_j \neq y_i$), we enforce a global repulsion across all modalities $\mathcal{M}_i = \{e_{I_i}, e_i^{\text{desc}}, e_i^{\text{syn}}, e_i^{\text{edge}}, e_i^{\text{neg}}\}$:
\begin{equation}
\mathcal{L}_{\text{inter}} = \sum_{r_i \in \mathcal{M}_i} \sum_{r_j \in \mathcal{M}_j} \max(0, \tau_{\text{inter}} - \|r_i - r_j\|_2).
\end{equation}

The overall Stage 1 alignment loss is defined as $\mathcal{L}_{\text{align}} = \lambda_{\text{intra}}\mathcal{L}_{\text{intra}} + \lambda_{\text{cross}}\mathcal{L}_{\text{cross}} + \lambda_{\text{inter}}\mathcal{L}_{\text{inter}}$, where $\lambda$'s balancing hyperparameters. This establishes a stable global semantic structure, enabling robust concept-level learning in Stage 2.

% ---------------------------------------------------------

\subsection{Concept-Level Alignment and Bottleneck Aggregation}
\label{sec:stage2}

Building upon the globally aligned semantic space, Stage 2 performs fine-grained concept extraction and filtering. To ensure robustness against linguistic sparsity and inter-cohort heterogeneity, this stage incorporates the Unified Concept Prototype Codebook (UCPC), Learnable Evidence Aggregation, and Image-Conditioned Reliability Gating (ICRG).

\noindent\textbf{Unified Concept Prototype Codebook (UCPC).} 
Unlike Stage 1's instance-level operation, UCPC is constructed strictly concept-wise. For each concept $c \in \mathcal{C}$ and state value $v \in \mathcal{V}_c$, clinical definitions may vary across datasets $d \in \{1, \dots, N\}$. Let $\mathcal{S}_{c,v}$ denote the subset of datasets containing annotations for $(c,v)$, and $\tilde{T}_{c,v,d}^x$ be the corresponding dataset-specific textual specifications ($x \in \{\text{desc}, \text{syn}, \text{edge}, \text{neg}\}$). To construct a cohort-agnostic representation, we extract and average the embeddings across all available datasets:
\begin{equation}
e_{c,v}^x = \frac{1}{|\mathcal{S}_{c,v}|} \sum_{d \in \mathcal{S}_{c,v}} \mathrm{Norm}\left(P_T(F_{\text{T}}(\tilde{T}_{c,v,d}^x))\right) \in \mathbb{R}^d.
\end{equation}
This operation effectively marginalizes out dataset-specific linguistic biases. The UCPC tensor for concept $c$ is defined as $\mathcal{T}_c = [e_{c,v}^x] \in \mathbb{R}^{(V_c \times 4) \times d}$, which is precomputed and frozen during inference.

To extract visual queries targeting these prototypes, we define $C$ learnable concept anchors $A \in \mathbb{R}^{C \times d_v}$, where $a_c = A[c,:]$ stands for the anchor for concept $c$. Given multi-scale patch tokens $F^{(\ell)}$ from the visual encoder, we perform per-layer cross-attention $E^{(\ell)} = \mathrm{LN}(\mathrm{FFN}(\mathrm{MHA}(Q^{(\ell)}, F^{(\ell)}, F^{(\ell)})) + Q^{(\ell)})$, where $Q^{(\ell)}$ is broadcasted from $A$, $FFN$ and $MHA$ stand for feed-forward network and multi-head attention. We aggregate multi-scale cues via learned weights $a_\ell$ as $E = \sum_{\ell=1}^L a_\ell E^{(\ell)}$. Finally, projection and normalization yield the visual queries $\hat{Q} = \mathrm{Norm}(\text{Proj.}(E)) \in \mathbb{R}^{C \times d}$, where $\hat{q}_c = \hat{Q}[c,:]$ probes concept $c$.

\noindent\textbf{Learnable Evidence Aggregation (LEA).} 
For each concept $c$, we compute the cosine similarity matrix between the visual query and the codebook slots as $S_c = \langle \hat{q}_c \cdot \mathcal{T}_c^\top \rangle \in \mathbb{R}^{V_c \times 4}$. We split the slots into positive evidence $\mathcal{P} = \{\text{desc}, \text{syn}, \text{edge}\}$ and negative evidence $\mathcal{N} = \{\text{neg}\}$. Utilizing learned slot weights $\alpha_c = \mathrm{softmax}(w_c) \in \mathbb{R}^4$ and a negative penalty scale $\beta_c = \mathrm{softplus}(b_c)$, the penalized score for value $v$ is calculated as:
\begin{equation}
s_c[v] = \sum_{x \in \mathcal{P}} \alpha_c[x] S_c[v,x] - \beta_c \sum_{x \in \mathcal{N}} \alpha_c[x] S_c[v,x].
\end{equation}
A linear calibration head $H_c$ refines these raw scores into $\hat{s}_c \in \mathbb{R}^{V_c}$, which are then concatenated across all concepts to form the dense concept bottleneck vector $z \in \mathbb{R}^{\sum_c V_c}$.

\noindent\textbf{Image-Conditioned Reliability Gating (ICRG).} 
To address concept sparsity across varying clinical cohorts, ICRG predicts concept applicability purely from visual features. A lightweight MLP $g(\cdot)$ takes pooled features $\bar{E} = \mathrm{meanpool}(E^{(L)})$ to output a reliability score $r = g(\bar{E}) \in (0,1)^C$. During training, $r$ is multiplied by a cohort-specific applicability mask $m_i \in \{0,1\}^C$ (expanded to match the $z$ dimension). At inference, the hard mask $m_i$ is discarded, yielding the final gated bottleneck $z'' = z \odot \mathrm{expand}(r, \{V_c\})$ that feeds into the classification head $\hat{y} = h(z'')$, where $\odot$ stands for element-wise multiplication.

\begin{table}[t!]
\centering
\caption{Dataset statistics including concept numbers and data splits.}
\label{tab:dataset_stats}
\resizebox{\linewidth}{!}{
\begin{tabular}{ccccc}
\toprule
\multirow{2}{*}{Datasets} & \multirow{2}{*}{Derm7pt~\cite{kawahara2018seven}} & \multirow{2}{*}{PH2~\cite{mendoncca2015ph2}} & \multicolumn{2}{c}{SkinCon~\cite{daneshjou2022skincon}} \\
\cmidrule(lr){4-5}
 &  &  & Fitzpatrick17k~\cite{groh2021evaluating} & DDI~\cite{daneshjou2022disparities} \\
\midrule
Concept  & 10 & 5 & 48 & 48 \\
Train set & 809 & - & 2952 & 524 \\
Validation set & 101 & - & 369 & 66 \\
Test set & 101 & 120 & 369 & 66 \\
Concept type & Dermoscopic & Dermoscopic & Clinical & Clinical \\
\bottomrule
\end{tabular}
}
\end{table}

\subsection{Three-Step Joint Optimization and Post-hoc Intervention}

To effectively decouple interpretable concept representation learning from the final diagnostic decision, our framework is optimized via a progressive three-step strategy. 

\noindent\textbf{Step 1: Concept Representation Alignment.}
We freeze the final classification head $h$ and exclusively optimize the upstream modules (encoders, projectors, learnable aggregation, and ICRG). Crucially, the concept prediction loss $\mathcal{L}_{\text{concept}}$ utilizes the cohort mask $m_i$ to ensure gradients only flow for concepts genuinely observable in the current sample's originating dataset:
\begin{equation}
\mathcal{L}_{\text{concept}} = \frac{1}{\sum_c m_{i,c}} \sum_{c} m_{i,c} \cdot \mathrm{BCE}(\hat{s}_{i,c}, c^*_{i,c}).
\end{equation}
Simultaneously, $m_i$ serves as the privileged target for the ICRG mechanism to learn applicability priors: $\mathcal{L}_{\text{gate}} = \mathrm{BCE}(r, m_i)$.
The objective for Step 1 integrates the Stage 1 alignment loss: $\mathcal{L}_{\text{step1}} = \lambda_{\text{align}} \mathcal{L}_{\text{align}} + \lambda_{\text{cpt}} \mathcal{L}_{\text{concept}} + \lambda_{\text{gate}} \mathcal{L}_{\text{gate}}$.

\noindent\textbf{Step 2: Diagnostic Head Initialization.}
We freeze all preceding modules and exclusively train the classification head $h$ using the filtered, high-confidence concept representations $z''$. Diagnosis classification is optimized using standard cross-entropy loss, ensuring the classifier maps pure semantic concepts to disease categories without overfitting to visual noise:
\begin{equation}
\mathcal{L}_{\text{cls}} = - \sum_{y \in \mathcal{Y}} y^* \log \hat{y}.
\end{equation}

\noindent\textbf{Step 3: End-to-End Fine-Tuning.}
Finally, we unfreeze the entire network (excluding the precomputed text representations in UCPC) and perform joint fine-tuning with a smaller learning rate. The comprehensive optimization objective is:
\begin{equation}
\mathcal{L}_{\text{total}} = \mathcal{L}_{\text{cls}} + \lambda_{\text{align}} \mathcal{L}_{\text{align}} + \lambda_{\text{cpt}} \mathcal{L}_{\text{concept}} + \lambda_{\text{gate}} \mathcal{L}_{\text{gate}}.
\end{equation}
where $\lambda$'s hyper-parameters balancing the multi-task objectives. 

\noindent\textbf{Multi-Modality Post-hoc Intervention.}
During inference, the UCPC and ICRG operate independently of cohort identifiers, establishing a transparent bottleneck $z''$. Unlike traditional dataset-isolated CBMs, our unified concept space intrinsically encompasses both macroscopic clinical ($\mathcal{C}_{\text{clin}}$) and microscopic dermoscopic ($\mathcal{C}_{\text{derm}}$) concepts, such that $\mathcal{C} = \mathcal{C}_{\text{clin}} \cup \mathcal{C}_{\text{derm}}$. 

When a clinician intervenes, he define an intervention set $\mathcal{I} \subseteq \mathcal{C}$. For each intervened concept $c \in \mathcal{I}$, the expert assigns a ground-truth state $v^*_c \in \mathcal{V}_c$. We construct the intervened concept representation $\tilde{s}_c$ via a piecewise override:
\begin{equation}
\tilde{s}_c = 
\begin{cases} 
\mathbf{o}(v^*_c), & \text{if } c \in \mathcal{I}, \\
\hat{s}_c, & \text{if } c \notin \mathcal{I}.
\end{cases}
\end{equation}
where $\mathbf{o}(v^*_c) \in \{0,1\}^{V_c}$ is the one-hot encoded vector of the assigned state. These locally modified representations are concatenated to form the updated bottleneck $\tilde{z} = \mathrm{concat}(\tilde{s}_1, \dots, \tilde{s}_C)$. The frozen classification head $h$ then dynamically recalculates the diagnosis as $\tilde{y} = h(\tilde{z})$.

Crucially, this formulation enables cross-modal corrections where the intervention set $\mathcal{I}$ bridges imaging types. Clinicians can apply dermoscopic concepts to macroscopic photographs (e.g., identifying arborizing vessels) and vice versa (e.g., detailing scales on dermoscopic inputs) when visual cues permit. This unique flexibility empowers clinicians to guide the model's decision path using holistic clinical expertise, regardless of the underlying imaging modality.

\begin{table}[t]
\centering
\caption{Diagnosis performance on Derm7pt~\cite{kawahara2018seven} 
and SkinCon~\cite{daneshjou2022skincon}.}
\label{tab:main_results_reformatted}
\small
\resizebox{\linewidth}{!}{
\begin{tabular}{c | l | c c c c c}
\toprule
Datasets & Methods & Precision & Recall & F1-score & AUROC & AUPRC \\
\midrule

\multirow{13}{*}{Derm7pt~\cite{kawahara2018seven}}
& CBM~\cite{koh2020concept}
& $0.774_{\pm 0.03}$ & $0.751_{\pm 0.02}$ & $0.762_{\pm 0.01}$ & $\textcolor{green!70!black}{0.885_{\pm 0.04}}$ & $0.752_{\pm 0.02}$ \\

& PCBM~\cite{yuksekgonul2022post}
& $0.784_{\pm 0.01}$ & $0.765_{\pm 0.03}$ & $0.774_{\pm 0.04}$ & $\textcolor{blue}{0.890_{\pm 0.02}}$ & $0.760_{\pm 0.01}$ \\

& PCBM-h~\cite{yuksekgonul2022post}
& $0.731_{\pm 0.02}$ & $0.720_{\pm 0.03}$ & $0.725_{\pm 0.01}$ & $0.875_{\pm 0.04}$ & $0.740_{\pm 0.02}$ \\

& LF-CBM~\cite{oikarinen2023label}
& $0.751_{\pm 0.03}$ & $0.740_{\pm 0.04}$ & $0.745_{\pm 0.02}$ & $0.880_{\pm 0.01}$ & $0.790_{\pm 0.01}$ \\

& LaBo~\cite{yang2023language}
& $\textcolor{blue}{0.823_{\pm 0.02}}$
& $0.775_{\pm 0.03}$
& $\textcolor{green!70!black}{0.798_{\pm 0.01}}$
& $0.836_{\pm 0.04}$
& $0.801_{\pm 0.02}$ \\

& CBI-VLM~\cite{patricio2024towards}
& $0.723_{\pm 0.01}$ & $0.730_{\pm 0.04}$ & $0.726_{\pm 0.02}$ & $0.865_{\pm 0.03}$ & $\textcolor{blue}{0.830_{\pm 0.04}}$ \\

& MICA~\cite{bie2024mica}
& $\textcolor{green!70!black}{0.822_{\pm 0.03}}$
& $\textcolor{blue}{0.785_{\pm 0.02}}$
& $\textcolor{blue}{0.802_{\pm 0.01}}$
& $0.828_{\pm 0.02}$
& $\textcolor{green!70!black}{0.811_{\pm 0.03}}$ \\

& Explicd~\cite{gao2024aligning}
& $0.811_{\pm 0.04}$
& $0.779_{\pm 0.01}$
& $0.795_{\pm 0.02}$
& $0.815_{\pm 0.03}$
& $0.809_{\pm 0.01}$ \\

& MAKE$^{*}$~\cite{yan2025make}
& $0.737_{\pm 0.02}$ & $0.729_{\pm 0.03}$ & $0.722_{\pm 0.01}$ & $0.882_{\pm 0.04}$ & $0.807_{\pm 0.02}$ \\

& TTI-CBM~\cite{he2025training}
& $0.794_{\pm 0.01}$
& $\textcolor{green!70!black}{0.780_{\pm 0.03}}$
& $0.787_{\pm 0.04}$
& $0.805_{\pm 0.02}$
& $0.772_{\pm 0.03}$ \\

& CoPA~\cite{dong2025copa}
& $0.723_{\pm 0.02}$ & $0.722_{\pm 0.01}$ & $0.717_{\pm 0.03}$ & $0.880_{\pm 0.02}$ & $0.806_{\pm 0.04}$ \\

& PanDerm$^{*}$~\cite{yan2025multimodal}
& $0.711_{\pm 0.03}$ & $0.714_{\pm 0.02}$ & $0.711_{\pm 0.01}$ & $0.843_{\pm 0.03}$ & $0.742_{\pm 0.02}$ \\

\cline{2-7}
& \textbf{UniCon (Ours)}
& $\textcolor{red}{\mathbf{0.845}_{\pm 0.02}}$
& $\textcolor{red}{\mathbf{0.824}_{\pm 0.01}}$
& $\textcolor{red}{\mathbf{0.828}_{\pm 0.04}}$
& $\textcolor{red}{\mathbf{0.903}_{\pm 0.03}}$
& $\textcolor{red}{\mathbf{0.844}_{\pm 0.01}}$ \\

\midrule

\multirow{13}{*}{SkinCon~\cite{daneshjou2022skincon}}
& CBM~\cite{koh2020concept}
& $0.795_{\pm 0.03}$ & $0.780_{\pm 0.02}$ & $0.787_{\pm 0.01}$ & $0.880_{\pm 0.04}$ & $0.845_{\pm 0.02}$ \\

& PCBM~\cite{yuksekgonul2022post}
& $0.815_{\pm 0.01}$ & $0.790_{\pm 0.03}$ & $0.802_{\pm 0.04}$ & $0.895_{\pm 0.02}$ & $0.865_{\pm 0.01}$ \\

& PCBM-h~\cite{yuksekgonul2022post}
& $0.800_{\pm 0.02}$ & $0.785_{\pm 0.01}$ & $0.792_{\pm 0.03}$ & $0.885_{\pm 0.04}$ & $0.850_{\pm 0.02}$ \\

& LF-CBM~\cite{oikarinen2023label}
& $0.805_{\pm 0.01}$ & $0.795_{\pm 0.04}$ & $0.800_{\pm 0.02}$ & $0.890_{\pm 0.03}$ & $0.860_{\pm 0.01}$ \\

& LaBo~\cite{yang2023language}
& $\textcolor{blue}{0.855_{\pm 0.02}}$
& $0.828_{\pm 0.03}$
& $\textcolor{blue}{0.841_{\pm 0.01}}$
& $0.932_{\pm 0.04}$
& $\textcolor{blue}{0.898_{\pm 0.03}}$ \\

& CBI-VLM~\cite{patricio2024towards}
& $0.810_{\pm 0.03}$ & $0.792_{\pm 0.01}$ & $0.801_{\pm 0.04}$ & $0.892_{\pm 0.02}$ & $0.855_{\pm 0.01}$ \\

& MICA~\cite{bie2024mica}
& $\textcolor{green!70!black}{0.850_{\pm 0.02}}$
& $0.822_{\pm 0.03}$
& $\textcolor{green!70!black}{0.836_{\pm 0.01}}$
& $0.925_{\pm 0.01}$
& $\textcolor{green!70!black}{0.890_{\pm 0.03}}$ \\

& Explicd~\cite{gao2024aligning}
& $0.842_{\pm 0.03}$ & $0.810_{\pm 0.01}$ & $0.826_{\pm 0.04}$ & $0.918_{\pm 0.01}$ & $0.885_{\pm 0.01}$ \\

& MAKE$^{*}$~\cite{yan2025make}
& $0.827_{\pm 0.02}$
& $\textcolor{blue}{0.836_{\pm 0.03}}$
& $0.829_{\pm 0.01}$
& $\textcolor{blue}{0.935_{\pm 0.04}}$
& $0.889_{\pm 0.02}$ \\

& TTI-CBM~\cite{he2025training}
& $0.830_{\pm 0.01}$ & $0.805_{\pm 0.04}$ & $0.817_{\pm 0.02}$ & $0.912_{\pm 0.02}$ & $0.875_{\pm 0.01}$ \\

& CoPA~\cite{dong2025copa}
& $0.816_{\pm 0.02}$
& $\textcolor{green!70!black}{0.834_{\pm 0.01}}$
& $0.824_{\pm 0.03}$
& $\textcolor{green!70!black}{0.935_{\pm 0.02}}$
& $\textcolor{green!70!black}{0.890_{\pm 0.03}}$ \\

& PanDerm$^{*}$~\cite{yan2025multimodal}
& $0.742_{\pm 0.01}$ & $0.808_{\pm 0.02}$ & $0.754_{\pm 0.03}$ & $0.897_{\pm 0.04}$ & $0.826_{\pm 0.02}$ \\

\cline{2-7}
& \textbf{UniCon (Ours)}
& $\textcolor{red}{\mathbf{0.861}_{\pm 0.03}}$
& $\textcolor{red}{\mathbf{0.870}_{\pm 0.02}}$
& $\textcolor{red}{\mathbf{0.854}_{\pm 0.01}}$
& $\textcolor{red}{\mathbf{0.950}_{\pm 0.04}}$
& $\textcolor{red}{\mathbf{0.918}_{\pm 0.02}}$ \\

\bottomrule
\end{tabular}}
\end{table}

\section{Experiments}

% \begin{wraptable}{r}{0.5\linewidth}
% \raggedright
% \caption{Statistics of datasets used in our experiments, including the number of concepts, data splits, and concept types. \\ De.= Dermoscopic, F17k= Fitzpatrick17k.}
% \label{tab:dataset_stats}
% \resizebox{\linewidth}{!}{
% \begin{tabular}{lcccccc}
% \toprule
% \multirow{2}{*}{\textbf{Datasets}} & \multicolumn{2}{c}{\multirow{2}{*}{Derm7pt\cite{kawahara2018seven}}} & \multicolumn{2}{c}{\multirow{2}{*}{PH2}} & \multicolumn{2}{c}{SkinCon} \\
%  & & & & & F17k & DDI \\
% \midrule
% \textbf{Concept N} & \multicolumn{2}{c}{10} & \multicolumn{2}{c}{5} & 48 & 48 \\
% \textbf{Train set} & \multicolumn{2}{c}{809} & \multicolumn{2}{c}{160} & 2952 & 524 \\
% \textbf{Valid set} & \multicolumn{2}{c}{101} & \multicolumn{2}{c}{20} & 369 & 66 \\
% \textbf{Test set} & \multicolumn{2}{c}{101} & \multicolumn{2}{c}{20} & 369 & 66 \\
% \textbf{Concept type} & \multicolumn{2}{c}{De.} & \multicolumn{2}{c}{De.} & \multicolumn{2}{c}{Clinical} \\
% \bottomrule
% \end{tabular}%
% }
 %\vspace{-1\baselineskip}
% \end{wraptable}

\begin{figure*}[!ht]
  \centering \includegraphics[width=0.96\linewidth]{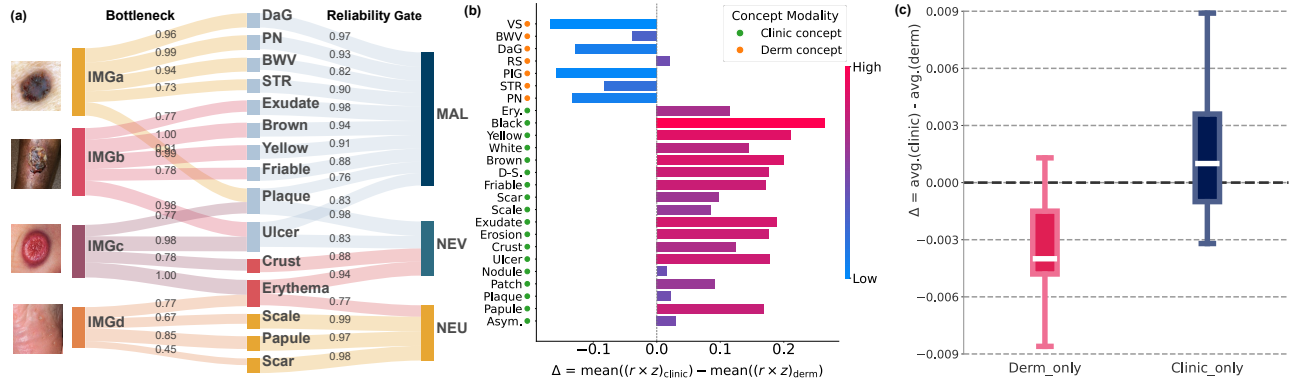}
  \caption{Concept bottleneck and reliability gate visualization. (a) Sankey flow from image evidence through gated concepts to diagnosis. (b) Test-set modality differences in reliability $r$ and bottleneck output $z$, with positive values for clinical concepts and mostly negative values for dermoscopic concepts. (c) Boxplots of $\Delta$ show clear separation between clinic-only and derm-only concepts.
  }
  \label{fig:gate_sanki}
\end{figure*}

% \subsection{Dataset and Implementation Details}
Our framework is evaluated on several public skin lesion datasets, including both dermoscopic datasets (PH2~\cite{mendoncca2015ph2} and Derm7pt~\cite{kawahara2018seven}) and a clinical dataset (SkinCon~\cite{daneshjou2022skincon} with two sub-cohorts: Fitzpatrick17k~\cite{groh2021evaluating} and DDI~\cite{daneshjou2022disparities}). More details are reported in Table~\ref{tab:dataset_stats}.
% \noindent\textbf{Dermoscopic Datasets:} \textbf{PH2}~\cite{mendoncca2015ph2} comprises 200 dermoscopic images annotated with 5 morphological concept features. By merging subcategories encompassing common and atypical nevi classes, the final dataset consists of 160 nevus and 40 melanoma cases. \textbf{Derm7pt}~\cite{kawahara2018seven} consists of 1,011 dermoscopic image cases. While the annotations are fundamentally based on the clinical 7-point checklist, our experiments utilize a total of 10 dermoscopic concepts.
% \noindent\textbf{Clinical Datasets:} We utilize the \textbf{SkinCon}~\cite{daneshjou2022skincon} dataset, which provides dense concept annotations for macroscopic clinical images. This encompasses two sub-cohorts: Fitzpatrick17k~\cite{groh2021evaluating}, containing 3,690 clinical images spanning non-neoplastic, benign, and malignant disease categories, and DDI~\cite{groh2021evaluating}, comprising 656 clinical images. 
 We cleaned the SkinCon datasets by removing sparse concepts with >99\% samples missing annotation values, retaining 18 valid clinical concepts. SkinCon and Derm7pt datasets are combined and partitioned into training, validation, and test sets as 8:1:1. PH2 is used as the external dataset for the purpose of zero-shot inference. All experiments are conducted on a workstation with 8 NVIDIA RTX 5090 GPUs.

% \setlength{\floatsep}{6pt}
% \textbf{Implementation Details.} Our proposed framework is systematically trained in two stages. During Stage 1 (global instance-level alignment), the network is optimized using the Adam optimizer with a learning rate of $1 \times 10^{-4}$ and a batch size of 512. In Stage 2 (concept-level alignment and diagnostic classification), we preserve the same Adam optimizer and $1 \times 10^{-4}$ learning rate, while the batch size is adjusted to 128. All experiments and model evaluations are conducted on a workstation equipped with a single NVIDIA RTX 5090 GPU.

\subsection{Comparison Results for Diagnosis}

Table \ref{tab:main_results_reformatted} compares our method with 12 state-of-the-art baselines on the Derm7pt and SkinCon datasets, where $^{*}$ denotes BioMedCLIP initialization without loading the official method-specific pretrained checkpoint. The top three results are highlighted in \textcolor{red}{\textbf{red}}, \textcolor{blue}{blue}, and \textcolor{green!70!black}{green}. As shown, UniCon consistently achieves the best performance across all evaluated metrics on both datasets. Notably, it secures the highest AUROC and AUPRC, which are critical threshold-independent metrics for highly imbalanced medical data. Furthermore, UniCon yields the highest F1-scores of 0.828 on Derm7pt and 0.854 on SkinCon, demonstrating superior overall diagnostic performance. While baselines like LaBo~\cite{yang2023language}, MICA~\cite{bie2024mica}, and MAKE~\cite{yan2025make} show competitive results as runners-up, our framework consistently dominates. This validates that UniCon successfully balances Precision and Recall without sacrificing cross-site generalization, providing highly accurate and interpretable diagnoses.

\subsection{Comparison Results for Concept Prediction}
For concept-based models, concept prediction accuracy directly dictates the reliability and interpretability of downstream disease diagnosis. Table~\ref{tab:concept_prediction} shows that our UniCon consistently ranks first in AUROC and AUPRC across all datasets (e.g., an AUROC of $0.910$ on Derm7pt). This confirms its robust capacity to activate rare but critical clinical concepts, ensuring the downstream classifier is driven by genuine evidence rather than spurious correlations. 

While some baselines peak in specific threshold-dependent metrics (e.g., MAKE\cite{yan2025make}'s Recall of $0.762$ on Derm7pt, or LaBo~\cite{yang2023language}'s Recall of $0.808$ on SkinCon), such extreme bias often introduces noisy false-positive concepts into the bottleneck. For instance, LaBo~\cite{yang2023language}'s high Recall explicitly sacrifices Precision ($0.830$), which can heavily confuse the final diagnosis. In contrast, UniCon achieves excellent Precision and strictly dominates the F1-scores ($0.778$ and $0.820$) across both datasets due to its balanced profile. This clean, informative bottleneck directly enhances the final diagnostic performance observed in Table~\ref{tab:main_results_reformatted}, resulting in a transparent and trustworthy clinical tool.

\begin{table}[htpb]
\centering
\caption{Concept prediction performance comparison on Derm7pt~\cite{kawahara2018seven} and SkinCon~\cite{daneshjou2022skincon} datasets.}
\label{tab:concept_prediction}
\small
\resizebox{\linewidth}{!}{
\begin{tabular}{c | l | c c c c c}
\toprule
Datasets & Methods & Precision & Recall & F1-score & AUROC & AUPRC \\
\midrule
\multirow{13}{*}{Derm7pt\cite{kawahara2018seven}} 
& CBM~\cite{koh2020concept} & $0.752_{\pm 0.08}$ & $0.725_{\pm 0.09}$ & $0.738_{\pm 0.08}$ & $0.861_{\pm 0.06}$ & $0.831_{\pm 0.08}$ \\
& PCBM~\cite{yuksekgonul2022post} & $0.760_{\pm 0.07}$ & $0.740_{\pm 0.08}$ & $0.750_{\pm 0.07}$ & $0.865_{\pm 0.07}$ & $0.838_{\pm 0.07}$ \\
& PCBM-h~\cite{yuksekgonul2022post} & $0.705_{\pm 0.09}$ & $0.695_{\pm 0.09}$ & $0.700_{\pm 0.09}$ & $0.850_{\pm 0.08}$ & $0.815_{\pm 0.09}$ \\
& LF-CBM~\cite{oikarinen2023label} & $0.728_{\pm 0.06}$ & $0.715_{\pm 0.07}$ & $0.721_{\pm 0.06}$ & $0.855_{\pm 0.06}$ & $0.828_{\pm 0.07}$ \\
& LaBo~\cite{yang2023language} & $\textcolor{green!70!black}{0.798}_{\pm 0.05}$ & $\textcolor{blue}{0.758}_{\pm 0.06}$ & $0.777_{\pm 0.05}$ & $\textcolor{green!70!black}{0.902}_{\pm 0.04}$ & $\textcolor{green!70!black}{0.868}_{\pm 0.05}$ \\
& CBI-VLM~\cite{patricio2024towards} & $0.701_{\pm 0.08}$ & $0.708_{\pm 0.07}$ & $0.704_{\pm 0.08}$ & $0.842_{\pm 0.07}$ & $0.809_{\pm 0.08}$ \\
& MICA~\cite{bie2024mica} & $0.795_{\pm 0.04}$ & $0.752_{\pm 0.05}$ & $\textcolor{blue}{0.773}_{\pm 0.04}$ & $0.904_{\pm 0.04}$ & $0.866_{\pm 0.04}$ \\
& Explicd~\cite{gao2024aligning} & $0.788_{\pm 0.05}$ & $0.750_{\pm 0.06}$ & $0.769_{\pm 0.05}$ & $0.891_{\pm 0.05}$ & $0.855_{\pm 0.05}$ \\
& MAKE$^{*}$~\cite{yan2025make} & $\textcolor{blue}{0.799}_{\pm 0.04}$ & $\textcolor{red}{\mathbf{0.762}}_{\pm 0.04}$ & $\textcolor{green!70!black}{0.780}_{\pm 0.04}$ & $0.901_{\pm 0.04}$ & $0.861_{\pm 0.05}$ \\
& TTI-CBM~\cite{he2025training} & $0.771_{\pm 0.06}$ & $0.751_{\pm 0.06}$ & $0.761_{\pm 0.06}$ & $0.880_{\pm 0.05}$ & $0.848_{\pm 0.06}$ \\
& CoPA~\cite{dong2025copa} & $0.793_{\pm 0.04}$ & $0.748_{\pm 0.05}$ & $0.770_{\pm 0.04}$ & $\textcolor{blue}{0.906}_{\pm 0.03}$ & $\textcolor{blue}{0.870}_{\pm 0.04}$ \\
& PanDerm$^{*}$~\cite{yan2025multimodal} & $0.786_{\pm 0.05}$ & $0.742_{\pm 0.06}$ & $0.763_{\pm 0.05}$ & $0.896_{\pm 0.04}$ & $0.863_{\pm 0.05}$ \\
\cline{2-7}
& \textbf{UniCon (Ours)} & $\textcolor{red}{\mathbf{0.802}}_{\pm 0.03}$ & $\textcolor{green!70!black}{0.755}_{\pm 0.04}$ & $\textcolor{red}{\mathbf{0.778}}_{\pm 0.03}$ & $\textcolor{red}{\mathbf{0.910}}_{\pm 0.02}$ & $\textcolor{red}{\mathbf{0.874}}_{\pm 0.03}$ \\
\midrule
\multirow{13}{*}{SkinCon~\cite{daneshjou2022skincon}} 
& CBM~\cite{koh2020concept} & $0.771_{\pm 0.07}$ & $0.755_{\pm 0.08}$ & $0.763_{\pm 0.07}$ & $0.855_{\pm 0.06}$ & $0.820_{\pm 0.07}$ \\
& PCBM~\cite{yuksekgonul2022post} & $0.790_{\pm 0.06}$ & $0.765_{\pm 0.07}$ & $0.777_{\pm 0.06}$ & $0.870_{\pm 0.06}$ & $0.840_{\pm 0.06}$ \\
& PCBM-h~\cite{yuksekgonul2022post} & $0.776_{\pm 0.08}$ & $0.760_{\pm 0.09}$ & $0.768_{\pm 0.08}$ & $0.860_{\pm 0.08}$ & $0.825_{\pm 0.08}$ \\
& LF-CBM~\cite{oikarinen2023label} & $0.780_{\pm 0.06}$ & $0.770_{\pm 0.07}$ & $0.775_{\pm 0.06}$ & $0.865_{\pm 0.06}$ & $0.835_{\pm 0.06}$ \\
& LaBo~\cite{yang2023language} & $\textcolor{green!70!black}{0.830}_{\pm 0.04}$ & $\textcolor{red}{\mathbf{0.808}}_{\pm 0.05}$ & $\textcolor{green!70!black}{0.819}_{\pm 0.04}$ & $\textcolor{blue}{0.907}_{\pm 0.04}$ & $\textcolor{blue}{0.872}_{\pm 0.04}$ \\
& CBI-VLM~\cite{patricio2024towards} & $0.785_{\pm 0.06}$ & $0.768_{\pm 0.07}$ & $0.776_{\pm 0.06}$ & $0.867_{\pm 0.06}$ & $0.830_{\pm 0.07}$ \\
& MICA~\cite{bie2024mica} & $0.825_{\pm 0.04}$ & $0.798_{\pm 0.05}$ & $0.811_{\pm 0.04}$ & $0.900_{\pm 0.04}$ & $0.865_{\pm 0.05}$ \\
& Explicd~\cite{gao2024aligning} & $0.818_{\pm 0.05}$ & $0.786_{\pm 0.06}$ & $0.802_{\pm 0.05}$ & $0.893_{\pm 0.05}$ & $0.860_{\pm 0.05}$ \\
& MAKE$^{*}$~\cite{yan2025make} & $0.821_{\pm 0.04}$ & $0.790_{\pm 0.05}$ & $0.805_{\pm 0.04}$ & $0.895_{\pm 0.04}$ & $0.863_{\pm 0.05}$ \\
& TTI-CBM~\cite{he2025training} & $0.805_{\pm 0.05}$ & $0.780_{\pm 0.06}$ & $0.792_{\pm 0.05}$ & $0.887_{\pm 0.05}$ & $0.850_{\pm 0.06}$ \\
& CoPA~\cite{dong2025copa} & $\textcolor{red}{\mathbf{0.837}}_{\pm 0.04}$ & $\textcolor{green!70!black}{0.802}_{\pm 0.05}$ & $\textcolor{blue}{0.819}_{\pm 0.04}$ & $\textcolor{green!70!black}{0.905}_{\pm 0.03}$ & $\textcolor{green!70!black}{0.870}_{\pm 0.04}$ \\
& PanDerm$^{*}$~\cite{yan2025multimodal} & $0.823_{\pm 0.05}$ & $0.800_{\pm 0.05}$ & $0.811_{\pm 0.05}$ & $0.903_{\pm 0.04}$ & $0.868_{\pm 0.05}$ \\
\cline{2-7}
& \textbf{UniCon (Ours)}& $\textcolor{blue}{0.835}_{\pm 0.03}$ & $\textcolor{blue}{0.806}_{\pm 0.03}$ & $\textcolor{red}{\mathbf{0.820}}_{\pm 0.02}$ & $\textcolor{red}{\mathbf{0.909}}_{\pm 0.02}$ & $\textcolor{red}{\mathbf{0.875}}_{\pm 0.03}$ \\
\bottomrule
\end{tabular}}
\end{table}
\subsection{Ablation Study Results}

\begin{table}[t]
\centering
\caption{Ablation study results on the SkinCon~\cite{daneshjou2022skincon} dataset.}
\label{tab:ablation}
\small
\resizebox{\linewidth}{!}{
\setlength{\tabcolsep}{3pt}
\begin{tabular}{c | l | c c c c c}
\toprule
Stage & Methods & Precision & Recall & F1-score & AUROC & AUPRC \\
\midrule

\multirow{7}{*}{1}
& w/o $T_i^{\text{desc}}$
& $0.828_{\pm 0.03}$ & $0.801_{\pm 0.02}$ & $0.785_{\pm 0.04}$ & $0.904_{\pm 0.01}$ & $0.869_{\pm 0.02}$ \\

& w/o $T_i^{\text{syn}}$
& $0.826_{\pm 0.04}$ & $0.799_{\pm 0.01}$ & $0.783_{\pm 0.03}$ & $0.902_{\pm 0.05}$ & $0.867_{\pm 0.02}$ \\

& w/o $T_i^{\text{edge}}$
& $0.824_{\pm 0.02}$ & $0.797_{\pm 0.04}$ & $0.781_{\pm 0.01}$ & $0.900_{\pm 0.03}$ & $0.865_{\pm 0.05}$ \\

& w/o $T_i^{\text{neg}}$
& $0.822_{\pm 0.05}$ & $0.795_{\pm 0.02}$ & $0.779_{\pm 0.03}$ & $0.898_{\pm 0.01}$ & $0.863_{\pm 0.04}$ \\

& w/o $T_i^{\text{desc}}, T_i^{\text{syn}}$
& $0.812_{\pm 0.01}$ & $0.785_{\pm 0.04}$ & $0.769_{\pm 0.02}$ & $0.888_{\pm 0.05}$ & $0.853_{\pm 0.03}$ \\

& w/o $T_i^{\text{desc}}, T_i^{\text{edge}}$
& $0.810_{\pm 0.02}$ & $0.783_{\pm 0.01}$ & $0.767_{\pm 0.04}$ & $0.886_{\pm 0.03}$ & $0.851_{\pm 0.05}$ \\

& w/o all MSS slots
& $0.798_{\pm 0.04}$ & $0.771_{\pm 0.05}$ & $0.755_{\pm 0.02}$ & $0.874_{\pm 0.01}$ & $0.839_{\pm 0.03}$ \\

\cmidrule(lr){1-7}

\multirow{7}{*}{2}
& single fixed template
& $0.825_{\pm 0.03}$ & $0.795_{\pm 0.02}$ & $0.782_{\pm 0.05}$ & $0.900_{\pm 0.04}$ & $0.862_{\pm 0.01}$ \\

& w/o $\tilde{T}^{\text{syn}}, \tilde{T}^{\text{edge}}$
& $0.838_{\pm 0.01}$ & $0.810_{\pm 0.05}$ & $0.808_{\pm 0.03}$ & $0.914_{\pm 0.02}$ & $0.878_{\pm 0.04}$ \\

& w/o $\tilde{T}^{\text{desc}}, \tilde{T}^{\text{syn}}$
& $0.836_{\pm 0.02}$ & $0.808_{\pm 0.04}$ & $0.793_{\pm 0.01}$ & $0.912_{\pm 0.05}$ & $0.876_{\pm 0.03}$ \\

& w/o $\tilde{T}^{\text{neg}}, \tilde{T}^{\text{edge}}$
& $0.834_{\pm 0.05}$ & $0.806_{\pm 0.03}$ & $0.791_{\pm 0.02}$ & $0.910_{\pm 0.04}$ & $0.874_{\pm 0.01}$ \\

& w/o $\tilde{T}^{\text{desc}}, \tilde{T}^{\text{neg}}$
& $0.846_{\pm 0.04}$ & $0.825_{\pm 0.01}$ & $0.805_{\pm 0.05}$ & $0.924_{\pm 0.03}$ & $0.890_{\pm 0.02}$ \\

& w/o $\tilde{T}^{\text{neg}}$
& $0.849_{\pm 0.03}$ & $0.818_{\pm 0.02}$ & $0.803_{\pm 0.04}$ & $0.928_{\pm 0.01}$ & $0.886_{\pm 0.05}$ \\

& w/o $r$
& $0.852_{\pm 0.02}$ & $0.822_{\pm 0.04}$ & $0.811_{\pm 0.05}$ & $0.936_{\pm 0.03}$ & $0.895_{\pm 0.01}$ \\

\cmidrule(lr){1-7}

\multirow{2}{*}{UniCon}
& solo-trained
& $0.847_{\pm 0.01}$ & $0.819_{\pm 0.03}$ & $0.806_{\pm 0.02}$ & $0.926_{\pm 0.04}$ & $0.888_{\pm 0.05}$ \\

& Ours
& $\mathbf{0.861}_{\pm 0.03}$
& $\mathbf{0.870}_{\pm 0.02}$
& $\mathbf{0.854}_{\pm 0.01}$
& $\mathbf{0.950}_{\pm 0.04}$
& $\mathbf{0.918}_{\pm 0.02}$ \\

\bottomrule
\end{tabular}}
\end{table}

Table~\ref{tab:ablation} details our ablation study on the SkinCon dataset. Removing Stage 1 text slots (\textit{w/o all MSS slots}) yields the lowest overall AUPRC (0.839). In Stage 2, a \textit{single fixed template} significantly underperforms the full model (AUROC 0.900 vs. 0.934), highlighting the necessity of multi-faceted features. Similarly, ablating fine-grained concept slots degrades discriminability; for instance, removing the negative counter-example slot (\textit{w/o} $\tilde{T}^{\text{neg}}$) drops Precision to 0.849. Crucially, our unified framework consistently outperforms the \textit{solo-trained} baseline, validating the advantage of a cross-cohort concept codebook over isolated training. Finally, omitting the reliability gating (\textit{w/o} $r$) decreases Precision (0.852), F1-score (0.811), and AUPRC (0.895), trading essential dynamic filtering for a negligible AUROC gain (0.936). Ultimately, the full framework delivers the most robust clinical performance.

\begin{figure}[!ht]
  \centering \includegraphics[width=\linewidth]{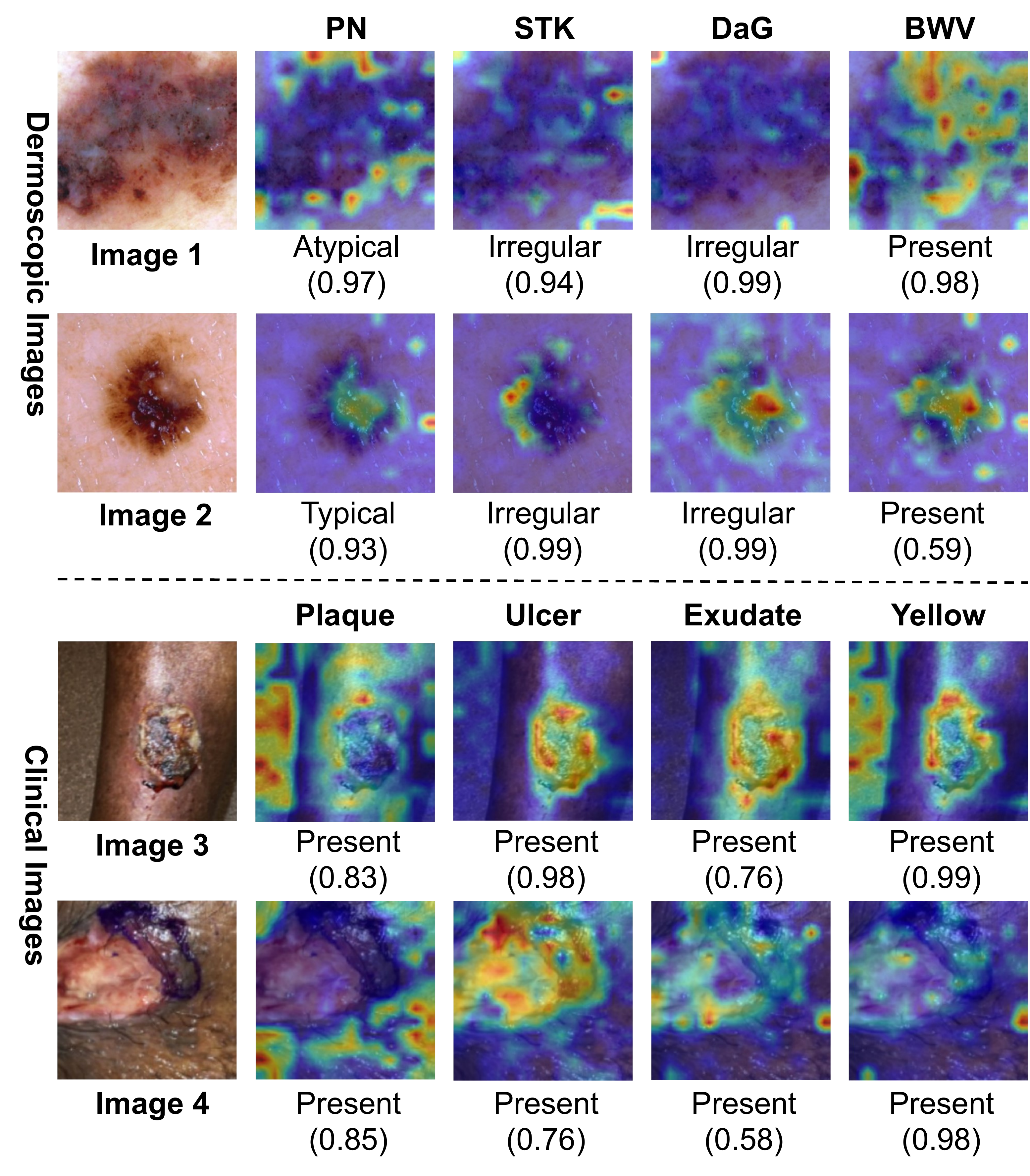}
  \caption{Grad-CAM visualization of dermoscopic and clinical images of our model. PN: pigment network; STK: streaks; DaG: dot and globals; BWV: blue-whitish veil.  
  }
  \label{fig:gradcam}
\end{figure}

\begin{figure}[!ht]
  \centering \includegraphics[width=0.94\linewidth]{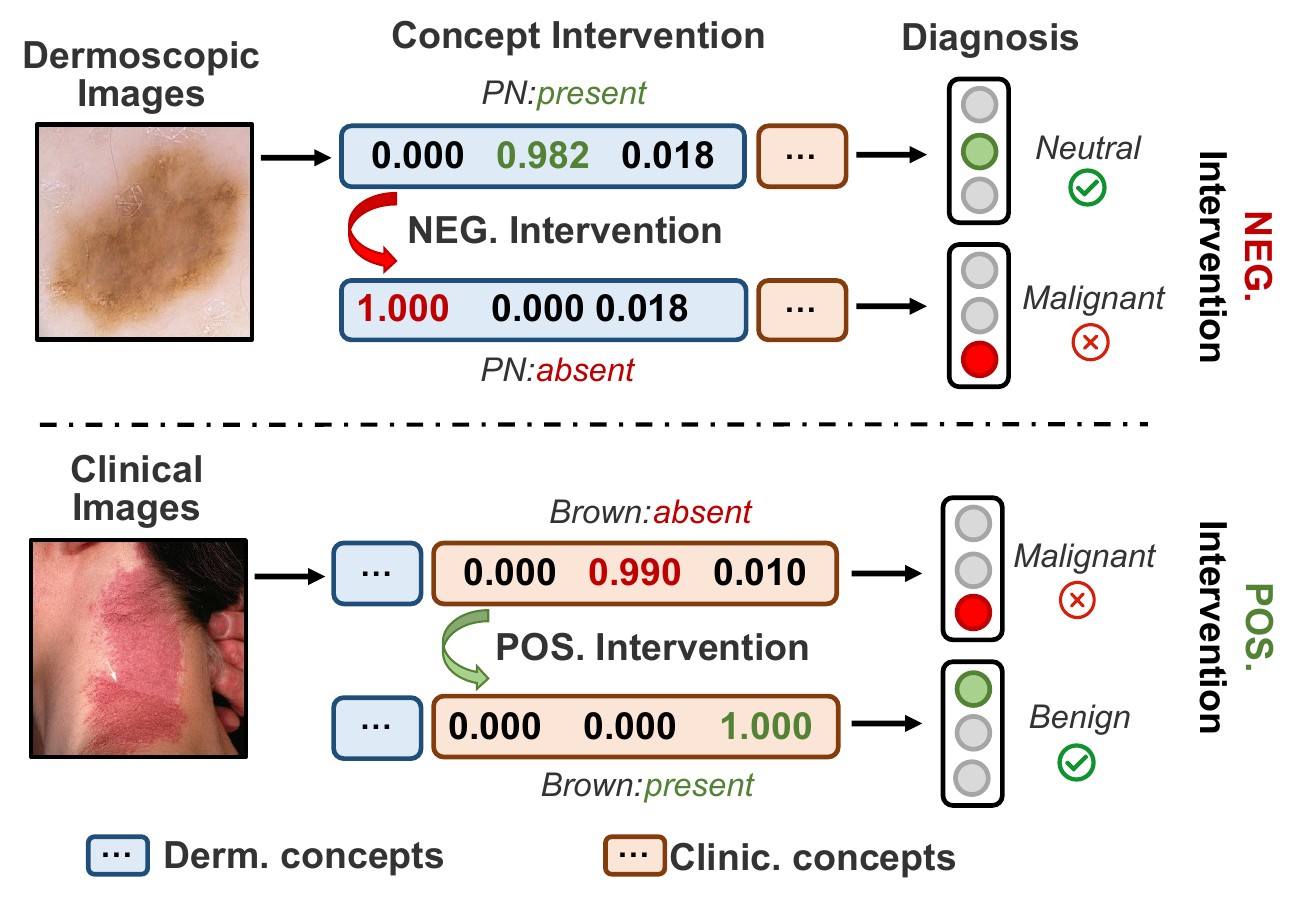 }
  \caption{Test-time positive and negative intervention visualization of dermoscopic and clinical images of our model.
  }
  \label{fig:intervention}
\end{figure}

% 单栏的bro
\begin{table}[t]
\centering
\caption{Zero-shot inference performance for disease diagnosis and concept prediction on the PH2~\cite{mendoncca2015ph2} dataset.}
\label{tab:zero_shot_ph2}
\resizebox{\linewidth}{!}{
\begin{tabular}{lccccc}
\toprule
\textbf{Method} & \textbf{Precision} & \textbf{Recall} & \textbf{F1-score} & \textbf{AUROC} & \textbf{AUPRC} \\
\midrule
\multicolumn{6}{c}{\textbf{Disease Diagnosis}} \\
\midrule
CBM~\cite{koh2020concept}       & $0.688$ & $0.675$ & $0.680$ & $0.728$ & $0.685$ \\
PCBM~\cite{yuksekgonul2022post}    & $0.701$ & $0.688$ & $0.695$ & $0.742$ & $0.698$ \\
PCBM-h~\cite{yuksekgonul2022post}  & $0.675$ & $0.658$ & $0.668$ & $0.715$ & $0.670$ \\
LF-CBM~\cite{oikarinen2023label}  & $0.710$ & $0.695$ & $0.702$ & $0.750$ & $0.705$ \\
LaBo~\cite{yang2023language}    & $0.735$ & $0.721$ & $\textcolor{blue}{0.735}$ & $\textcolor{blue}{0.781}$ & $\textcolor{green!70!black}{0.744}$ \\
Explicd~\cite{gao2024aligning} & $0.720$ & $0.711$ & $0.715$ & $\textcolor{green!70!black}{0.772}$ & $0.725$ \\
MAKE$^{*}$~\cite{yan2025make}    & $\textcolor{blue}{0.745}$ & $0.718$ & $\textcolor{green!70!black}{0.729}$ & $0.768$ & $\textcolor{blue}{0.750}$ \\
TTI-CBM~\cite{he2025training} & $0.715$ & $0.702$ & $0.708$ & $0.755$ & $0.712$ \\
CoPA~\cite{dong2025copa}    & $\textcolor{green!70!black}{0.738}$ & $\textcolor{red}{\mathbf{0.732}}$ & $0.725$ & $0.765$ & $0.739$ \\
\cline{1-6}
\textbf{UniCon (Ours)}  & $\textcolor{red}{\mathbf{0.751}}$ & $\textcolor{blue}{0.727}$ & $\textcolor{red}{\mathbf{0.739}}$ & $\textcolor{red}{\mathbf{0.786}}$ & $\textcolor{red}{\mathbf{0.754}}$ \\
\midrule
\multicolumn{6}{c}{\textbf{Concept Prediction}} \\
\midrule
CBM~\cite{koh2020concept}       & $0.688$ & $0.678$ & $0.682$ & $0.725$ & $0.684$ \\
PCBM~\cite{yuksekgonul2022post}    & $0.705$ & $0.692$ & $0.698$ & $0.738$ & $0.699$ \\
PCBM-h~\cite{yuksekgonul2022post}  & $0.672$ & $0.655$ & $0.665$ & $0.712$ & $0.668$ \\
LF-CBM~\cite{oikarinen2023label}  & $0.712$ & $0.701$ & $0.705$ & $0.745$ & $0.708$ \\
LaBo~\cite{yang2023language}    & $\textcolor{red}{\mathbf{0.751}}$ & $0.715$ & $\textcolor{green!70!black}{0.726}$ & $\textcolor{green!70!black}{0.772}$ & $\textcolor{blue}{0.742}$ \\
Explicd~\cite{gao2024aligning} & $0.725$ & $\textcolor{green!70!black}{0.720}$ & $0.718$ & $0.758$ & $0.722$ \\
MAKE$^{*}$~\cite{yan2025make}    & $0.732$ & $0.718$ & $\textcolor{blue}{0.732}$ & $\textcolor{red}{\mathbf{0.785}}$ & $\textcolor{green!70!black}{0.736}$ \\
TTI-CBM~\cite{he2025training} & $0.718$ & $0.708$ & $0.712$ & $0.752$ & $0.715$ \\
CoPA~\cite{dong2025copa}    & $\textcolor{green!70!black}{0.738}$ & $\textcolor{blue}{0.728}$ & $0.721$ & $0.765$ & $0.731$ \\
\cline{1-6}
\textbf{UniCon (Ours)}  & $\textcolor{blue}{0.744}$ & $\textcolor{red}{\mathbf{0.734}}$ & $\textcolor{red}{\mathbf{0.739}}$ & $\textcolor{blue}{0.779}$ & $\textcolor{red}{\mathbf{0.749}}$ \\
\bottomrule
\end{tabular}
}
\end{table}

\subsection{Cross-modal Retrieval Results}

Table~\ref{tab:cross-retreval} shows that our proposed UniCon achieves the best cross-modal retrieval performance on SkinCon (AVG 0.279). It significantly outperforms vision-language models like MAKE~\cite{yan2025make} (0.235) and the domain-specific BioMedCLIP~\cite{zhang2023biomedclip} (0.143) across both T2I and I2T tasks, highlighting the effectiveness of multi-faceted semantic specifications in bridging the modality gap. To evaluate robustness against clinical annotation noise, we introduce UniCon$^\dagger$, trained with 10\% randomly shuffled image-text pairs in Stage 1. Despite this severe perturbation, UniCon$^\dagger$ maintains strong performance (AVG 0.190), demonstrating that our unified representation space provides exceptional fault tolerance against mismatched data.

\subsection{Visualization}
\subsubsection{Explainable Results}
To verify the spatial grounding of the learned concepts, we employ Grad-CAM~\cite{selvaraju2017grad} visualizations (Fig.~\ref{fig:gradcam}). The results confirm that the proposed model captures concept-specific, localized evidence across both modalities. In dermoscopic cases, the model consistently attends to lesion subregions corresponding to morphological concepts (e.g., pigment network, streaks, and blue-whitish veil). In clinical cases, the highlighted regions align precisely with lesion-level semantics (e.g., plaque and ulcer). By extracting modality-specific representations from diagnostically relevant areas rather than relying on spurious global cues, these qualitative results demonstrate the effectiveness and transferability of the learned concept bottleneck across heterogeneous image modalities.

\subsubsection{Bottleneck and Reliability Gate Analysis}
Fig.~\ref{fig:gate_sanki} elucidates the inference dynamics of the concept bottleneck and reliability gate. The Sankey diagram (Fig.~\ref{fig:gate_sanki}(a)) shows a sparse, structured decision path, indicating that predictions are based on explainable concepts rather than direct correlations. Modality-wise differences in gated concept responses ($\Delta$), shown in Fig.~\ref{fig:gate_sanki}(b), reveal that the reliability gate amplifies modality-consistent concepts while suppressing mismatched ones, with positive $\Delta$ for clinical concepts and negative $\Delta$ for dermoscopic concepts. This modality-aware discrimination is statistically supported by distinctly separated $\Delta$ distributions for clinical-only and dermoscopic-only concepts (Fig.~\ref{fig:gate_sanki} (c)). In summary, the intermediate concept layer enables both interpretability and robust adaptation to heterogeneous image modalities.

\begin{figure}[!ht]
  \centering \includegraphics[width=\linewidth]{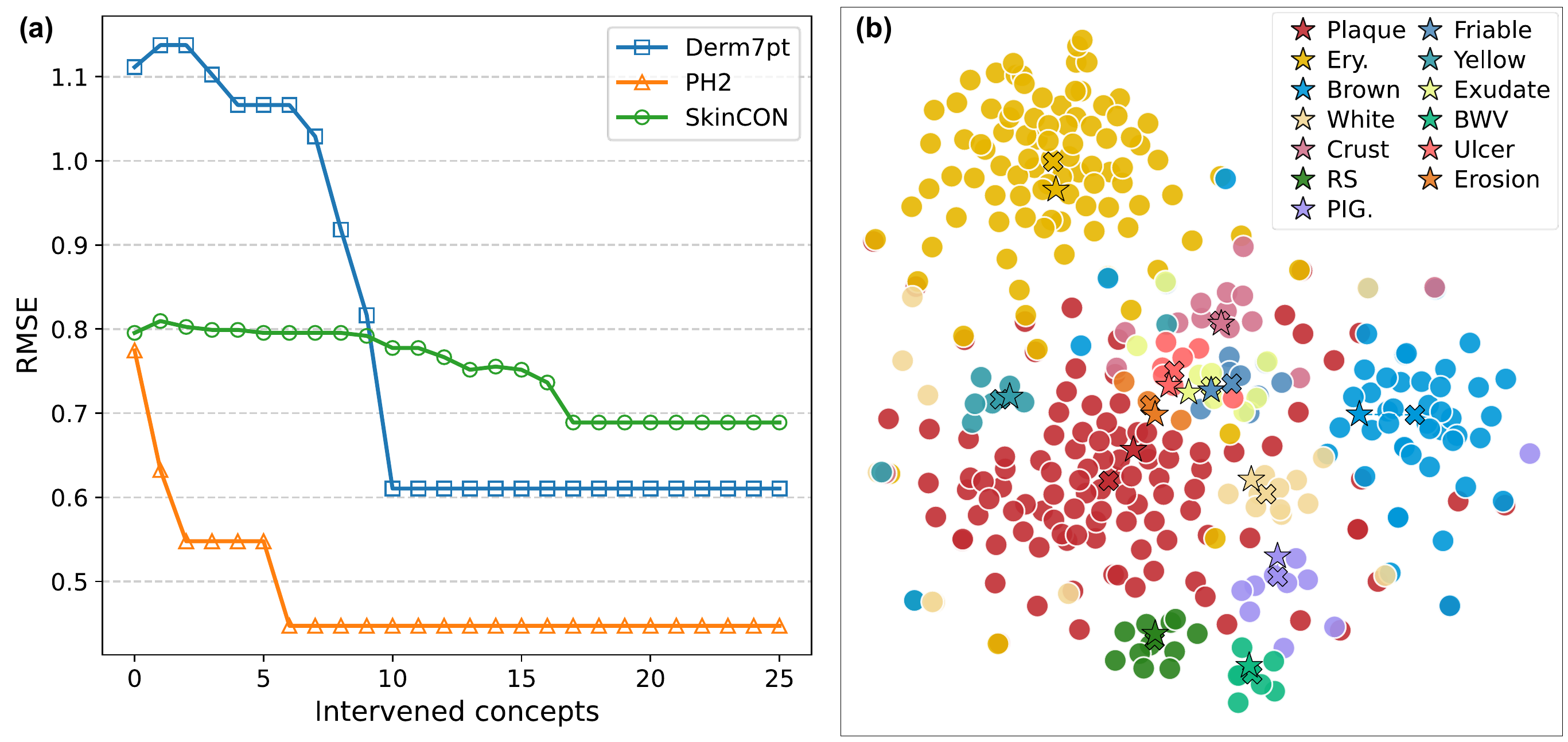 }
  \caption{Test-time intervention results of our model. (a) effective of test-time intervention. (b) t-SNE plot for different concepts and their prototypes.
  }
  \label{fig:intervention_results_fig}
\end{figure}

\subsection{Test-time Intervention Results}

% \begin{table}[ht]
% \centering
% \caption{Impact of post-hoc concept intervention on diagnostic performance on the PH2 dataset.}
% \label{tab:intervention}
% \resizebox{0.85\linewidth}{!}{
% \begin{tabular}{llcccc}
% \toprule
% & \textbf{\# ITV} & \textbf{ACC} & \textbf{$\Delta$ACC} & \textbf{RMSE} & \textbf{$\Delta$RMSE} \\
% \midrule
% ITV-Free & 0 & 73.9 & - & 0.882 & - \\
% \midrule
% \multirow{5}{*}{Neg ITV} 
% % & 3  & 73.5 & \textcolor{blue}{-0.4} & 0.891 & \textcolor{blue}{+0.009} \\
% & 5  & 73.1 & \textcolor{blue}{-0.8} & 0.893 & \textcolor{blue}{+0.012} \\
% & 7  & 72.3 & \textcolor{blue}{-1.7} & 0.919 & \textcolor{blue}{+0.037} \\
% & 10 & 71.0 & \textcolor{blue}{-2.9} & 0.946 & \textcolor{blue}{+0.064} \\
% & 15 & 71.4 & \textcolor{blue}{-2.5} & 0.944 & \textcolor{blue}{+0.062} \\
% \midrule
% \multirow{5}{*}{Pos ITV} 
% % & 3  & 76.5 & \textcolor{red}{+2.5} & 0.853 & \textcolor{red}{-0.029} \\
% & 5  & 76.9 & \textcolor{red}{+2.9} & 0.843 & \textcolor{red}{-0.039} \\
% & 7  & 78.2 & \textcolor{red}{+4.2} & 0.828 & \textcolor{red}{-0.054} \\
% & 10 & 79.8 & \textcolor{red}{+5.9} & 0.802 & \textcolor{red}{-0.080} \\
% & 15 & 81.1 & \textcolor{red}{+7.1} & 0.753 & \textcolor{red}{-0.129} \\
% \bottomrule
% \end{tabular}}
% \end{table}

\subsubsection{Unified Intervention Interface}
Fig.~\ref{fig:intervention} demonstrates the clinical utility of our unified, modality-agnostic test-time intervention framework. Modifying a single concept state directly and predictably alters the final diagnosis via interpretable feedback consistently across both image types. Quantitative results in Fig.~\ref{fig:intervention_results_fig} (a) and Table A1 in Appendix show that Root Mean Square Error(RMSE) steadily decreases as the number of intervened concepts increases. Specifically, providing ground-truth concepts (Pos ITV) enhances accuracy by up to 7.1\%, while injecting incorrect concepts (Neg ITV) proportionally degrades performance. This explicit responsiveness demonstrates our model's potential for reliable and controllable clinical use across diverse environments.

\subsubsection{Prototype Codebook Analysis}
The t-SNE~\cite{van2008visualizing} visualization, as shown in Fig.~\ref{fig:intervention_results_fig}(b), demonstrates that the learned concept-specific features form well-structured clusters in the embedding space. Notably, the learned unified concept prototypes (stars) closely align with the empirical cluster centers (crosses) of the projected test image features for each concept. This strong prototype-feature alignment confirms that the unified prototypes faithfully capture the semantic core of the representations, facilitating stable and interpretable concept modeling across diverse samples.

\begin{table}[htpb]
    \centering
    \caption{Cross-modal retrieval performance comparison on the SkinCon dataset~\cite{daneshjou2022skincon}. $\dagger$ stands for the method trained with 10\% random shuffle.}
    \resizebox{0.9\linewidth}{!}{
    \begin{tabular}{c|cc|cc|c}
\hline
\multirow{2}{*}{\textbf{Method}} & \multicolumn{2}{c|}{\textbf{I2T}} & \multicolumn{2}{c|}{\textbf{T2I}} & \multirow{2}{*}{\textbf{AVG}} \\
 & R@10 & R@50 & R@10 & R@50 & \\
\hline
MAKE$^{*}$~\cite{yan2025make} & \textcolor{red}{0.196} & \textcolor{blue}{0.382} & \textcolor{blue}{0.121} & 0.241 & \textcolor{blue}{0.235} \\
BioMedCLIP~\cite{zhang2023biomedclip} & 0.113 & 0.210 & 0.074 & 0.174 & 0.143 \\
UniCon$^\dagger$ (Ours) & 0.105 & 0.274 & 0.098 & \textcolor{blue}{0.284} & 0.190 \\
UniCon (Ours) & \textcolor{blue}{0.194} & \textcolor{red}{0.389} & \textcolor{red}{0.138} & \textcolor{red}{0.393} & \textcolor{red}{0.279} \\
\hline
\end{tabular}}
    \label{tab:cross-retreval}
\end{table}

\subsection{Zero-shot Inference Results}
To assess out-of-distribution robustness, we evaluate zero-shot inference on the unseen PH2 dataset (Table~\ref{tab:zero_shot_ph2}). Across both disease diagnosis and concept prediction tasks, our proposed UniCon consistently surpasses state-of-the-art baselines (e.g., PanDerm~\cite{yan2025multimodal}, LaBo, and MAKE). This superior diagnostic transferability is fundamentally driven by UniCon's precise identification of morphological and clinical concepts without prior target-domain exposure. Ultimately, these results confirm that UniCon captures robust, universally applicable semantic representations rather than overfitting to the source distribution, enabling reliable and interpretable generalization in heterogeneous clinical settings.

\section{Conclusion}

% We address cross-site concept heterogeneity in dermatological diagnosis, where variations in terminology, granularity, and annotation hinder scalable deployment and clinician-in-the-loop corrections. To overcome this, we propose UniCon, a unified multimodal framework that aligns heterogeneous concept systems, enriches semantic descriptions, and selects reliable image-conditioned evidence, enabling interpretable and transferable interventions. Experiments on multiple skin lesion datasets show improved diagnostic accuracy and robust test-time corrections. Future work will extend UniCon to other medical imaging domains, discover latent disease subtypes, and integrate dynamic knowledge graphs for disease progression modeling.

In this paper, we addressed the challenge of cross-site concept heterogeneity in dermatology image diagnosis, which stems from variations in terminology, semantic granularity, and annotation protocols. Prior approaches—including traditional concept-based models or concept-based vision-language models, relied heavily on dataset-specific engineering and lacked flexible intervention mechanisms, thereby limiting their scalable clinical deployment. To overcome these barriers, we proposed UniCon, a unified multimodal framework that successfully aligned heterogeneous concept taxonomies into a shared semantic interface via cross-site concept learning. By leveraging enriched textual specifications and a self-gating mechanism to select reliable, image-conditioned evidence, our framework enabled consistent, interpretable, and transferable concept-level interventions. UniCon enhances diagnostic accuracy and enables robust test-time corrections in skin lesion datasets, facilitating scalable, multi-center clinical applications.

\clearpage  % TODO FINAL: This \clearpage needs to be removed from both review and camera-ready versions.

\section{Acknowledgments}
This work was supported by Zhejiang Leading Innovative and Entrepreneur Team Introduction Program (No. 2024R01007), The “Pioneer” and “Leading Goose” Research and Development Program of Zhejiang (No. 2025C02077).

\bibliographystyle{ACM-Reference-Format}
\balance
\bibliography{sample-base}

\section{Appendix}
\subsection{Test-time Intervention Details}

\begin{table}[htbp]
\centering
\renewcommand{\arraystretch}{1.1} % 稍微增加行间距以提升可读性
\caption*{Table A1: Accuracy and RMSE changes of test-time concept intervention for dataset PH2.}
\begin{tabular}{lcccccc}
\toprule
& \textbf{\# ITV} & \textbf{ACC} & \textbf{$\Delta$ACC} & \textbf{Error} & \textbf{RMSE} & \textbf{$\Delta$RMSE} \\
\midrule
ITV-Free & 0 & 73.9 & - & 26.1 & 0.882 & - \\
\midrule
\multirow{18}{*}{Neg ITV} 
& 1  & 73.9 & \textcolor{blue}{0.0}  & 26.1 & 0.882 & \textcolor{blue}{0.000} \\
& 2  & 73.5 & \textcolor{blue}{-0.4} & 26.5 & 0.891 & \textcolor{blue}{0.009} \\
& 3  & 73.5 & \textcolor{blue}{-0.4} & 26.5 & 0.891 & \textcolor{blue}{0.009} \\
& 4  & 73.1 & \textcolor{blue}{-0.8} & 26.9 & 0.893 & \textcolor{blue}{0.012} \\
& 5  & 73.1 & \textcolor{blue}{-0.8} & 26.9 & 0.893 & \textcolor{blue}{0.012} \\
& 6  & 72.7 & \textcolor{blue}{-1.3} & 27.3 & 0.910 & \textcolor{blue}{0.028} \\
& 7  & 72.3 & \textcolor{blue}{-1.7} & 27.7 & 0.919 & \textcolor{blue}{0.037} \\
& 8  & 71.4 & \textcolor{blue}{-2.5} & 28.6 & 0.937 & \textcolor{blue}{0.055} \\
& 9  & 70.6 & \textcolor{blue}{-3.4} & 29.4 & 0.955 & \textcolor{blue}{0.073} \\
& 10 & 71.0 & \textcolor{blue}{-2.9} & 29.0 & 0.946 & \textcolor{blue}{0.064} \\
& 11 & 70.6 & \textcolor{blue}{-3.4} & 29.4 & 0.955 & \textcolor{blue}{0.073} \\
& 12 & 71.0 & \textcolor{blue}{-2.9} & 29.0 & 0.946 & \textcolor{blue}{0.064} \\
& 13 & 71.4 & \textcolor{blue}{-2.5} & 28.6 & 0.944 & \textcolor{blue}{0.062} \\
& 14 & 71.0 & \textcolor{blue}{-2.9} & 29.0 & 0.946 & \textcolor{blue}{0.064} \\
& 15 & 71.4 & \textcolor{blue}{-2.5} & 28.6 & 0.944 & \textcolor{blue}{0.062} \\
& 16 & 73.1 & \textcolor{blue}{-0.8} & 26.9 & 0.907 & \textcolor{blue}{0.026} \\
& 17 & 72.3 & \textcolor{blue}{-1.7} & 27.7 & 0.933 & \textcolor{blue}{0.051} \\
& 18 & 72.3 & \textcolor{blue}{-1.7} & 27.7 & 0.933 & \textcolor{blue}{0.051} \\
\midrule
\multirow{18}{*}{Pos ITV} 
& 1  & 75.6 & \textcolor{red}{1.7} & 24.4 & 0.850 & \textcolor{red}{-0.032} \\
& 2  & 76.5 & \textcolor{red}{2.5} & 23.5 & 0.853 & \textcolor{red}{-0.029} \\
& 3  & 76.5 & \textcolor{red}{2.5} & 23.5 & 0.853 & \textcolor{red}{-0.029} \\
& 4  & 76.5 & \textcolor{red}{2.5} & 23.5 & 0.853 & \textcolor{red}{-0.029} \\
& 5  & 76.9 & \textcolor{red}{2.9} & 23.1 & 0.843 & \textcolor{red}{-0.039} \\
& 6  & 76.9 & \textcolor{red}{2.9} & 23.1 & 0.843 & \textcolor{red}{-0.039} \\
& 7  & 78.2 & \textcolor{red}{4.2} & 21.8 & 0.828 & \textcolor{red}{-0.054} \\
& 8  & 78.2 & \textcolor{red}{4.2} & 21.8 & 0.828 & \textcolor{red}{-0.054} \\
& 9  & 78.6 & \textcolor{red}{4.6} & 21.4 & 0.817 & \textcolor{red}{-0.064} \\
& 10 & 79.8 & \textcolor{red}{5.9} & 20.2 & 0.802 & \textcolor{red}{-0.080} \\
& 11 & 79.8 & \textcolor{red}{5.9} & 20.2 & 0.802 & \textcolor{red}{-0.080} \\
& 12 & 79.8 & \textcolor{red}{5.9} & 20.2 & 0.794 & \textcolor{red}{-0.088} \\
& 13 & 80.7 & \textcolor{red}{6.7} & 19.3 & 0.772 & \textcolor{red}{-0.109} \\
& 14 & 80.7 & \textcolor{red}{6.7} & 19.3 & 0.764 & \textcolor{red}{-0.117} \\
& 15 & 81.1 & \textcolor{red}{7.1} & 18.9 & 0.753 & \textcolor{red}{-0.129} \\
& 16 & 81.5 & \textcolor{red}{7.6} & 18.5 & 0.742 & \textcolor{red}{-0.140} \\
& 17 & 82.8 & \textcolor{red}{8.8} & 17.2 & 0.716 & \textcolor{red}{-0.166} \\
& 18 & 82.8 & \textcolor{red}{8.8} & 17.2 & 0.716 & \textcolor{red}{-0.166} \\
\bottomrule
\end{tabular}
\end{table}

To evaluate the interpretability and intervention ability of our framework, we conduct test-time concept interventions, as detailed in Table A1. The results demonstrate a strict causal relationship between the intermediate concept states and the final diagnostic predictions. When applying positive interventions (Pos ITV)—replacing predicted concepts with expert-validated ground-truth labels—the model exhibits a consistent and monotonic performance improvement. Specifically, correcting 18 concepts yields a significant absolute accuracy gain of 8.8\% (reaching 82.8\%) and reduces the RMSE by 0.166. This highlights the framework's capability to seamlessly integrate human-in-the-loop feedback for refined clinical decision-making. Conversely, negative interventions (Neg ITV)—injecting incorrect concept states—proportionally degrade the diagnostic performance, increasing the RMSE by up to 0.073. This symmetric responsiveness explicitly verifies that our architecture does not bypass the bottleneck via shortcut learning; rather, the final diagnosis is genuinely driven by the transparent semantic concepts, ensuring a trustworthy and controllable diagnostic process.

\subsection{Hyperparameters Setting}

\begin{table*}[htbp]
\centering
\caption*{Table A2: Comprehensive hyperparameter configurations for the two-stage training process of UniCon.}
\label{tab:hyperparameters}
\resizebox{\linewidth}{!}{
\begin{tabular}{clccccc p{4.5cm}}
\toprule
\textbf{Stage} & \textbf{Training Phase} & \textbf{Optimizer} & \textbf{LR} & \textbf{WD} & \textbf{BS} & \textbf{Epochs} & \textbf{Loss Weights \& Margins} \\
\midrule
\textbf{Stage 1} & Instance-level Semantic Alignment & AdamW & $1 \times 10^{-4}$ & $1 \times 10^{-4}$ & 128 & 150 & 
$\lambda_{intra}=1.0$, $\lambda_{cross}=1.0$, $\lambda_{inter}=1.0$ \newline
$\tau_{close}=0.1$, $\tau_{spread}=0.9$ \newline
$\tau_{in}=0.2$, $\tau_{away}=0.8$ \newline
$\tau_{inter}=0.5$ \\
\midrule
\multirow{3}{*}{\textbf{Stage 2}} 
& Step 1: Concept Representation Alignment & AdamW & $1 \times 10^{-4}$ & $1 \times 10^{-4}$ & 64 & 100 & 
$\lambda_{align}=1.0$, $\lambda_{cpt}=1.0$, $\lambda_{gate}=0.5$ \\
\cmidrule{2-8}
& Step 2: Diagnostic Head Initialization & AdamW & $5 \times 10^{-4}$ & $1 \times 10^{-4}$ & 64 & 100 & 
Cross-Entropy Loss only ($\mathcal{L}_{cls}$) \\
\cmidrule{2-8}
& Step 3: End-to-End Fine-Tuning & AdamW & $1 \times 10^{-5}$ & $1 \times 10^{-4}$ & 64 & 50 & 
$\lambda_{align}=1.0$, $\lambda_{cpt}=1.0$, $\lambda_{gate}=0.5$ \\
\bottomrule
\end{tabular}
}
\end{table*}

\begin{figure*}[!tb]
  \centering \includegraphics[width=\linewidth]{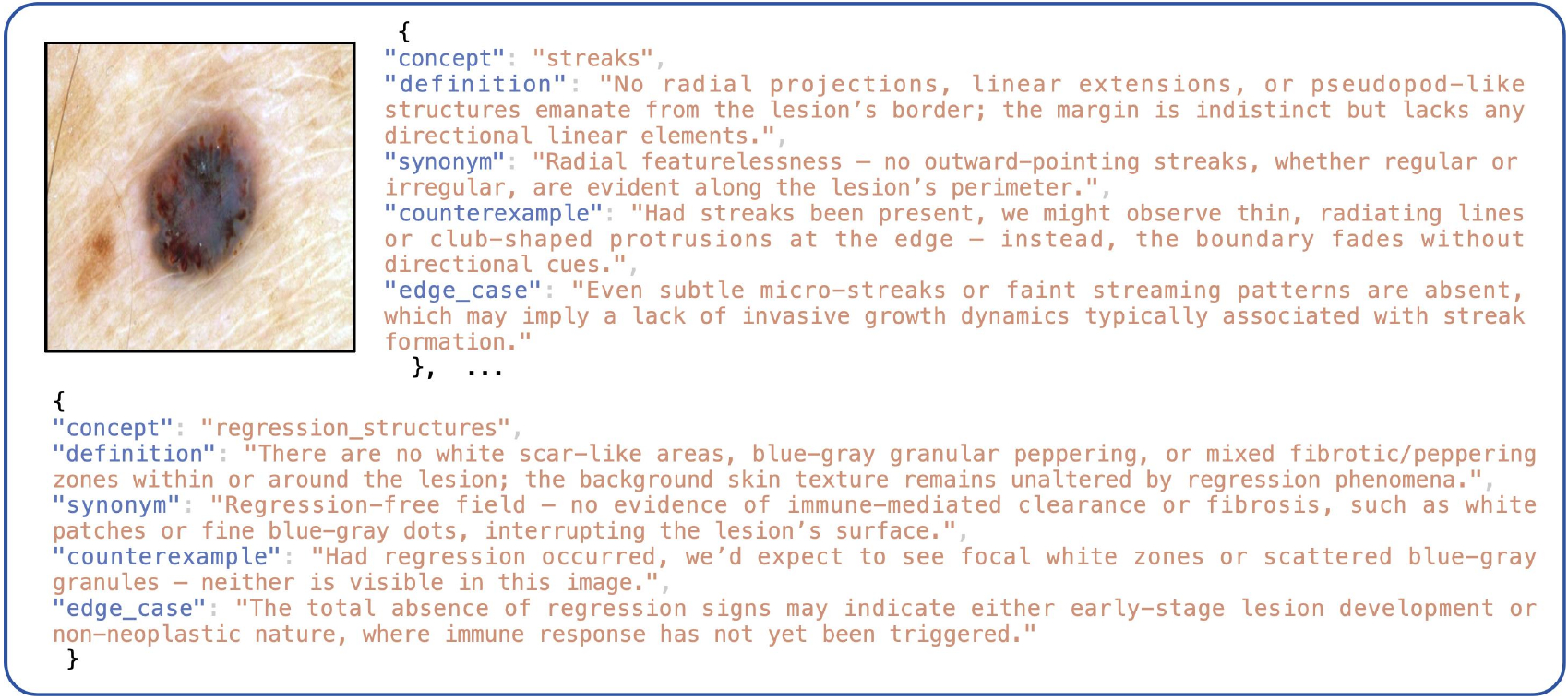}
  \caption*{Figure A1: Qualitative examples of the generated Multi-Faceted Semantic Specifications (MSS). The LVLM successfully outputs image-grounded, four-slot linguistic descriptions (Definition, Synonym, Counter-example, and Edge-cases) that explicitly detail instance-specific manifestations or absences.}
  % \vspace{-0.3cm}
  \label{fig:example}
\end{figure*}

\begin{figure*}[!tb]
  \centering \includegraphics[width=\linewidth]{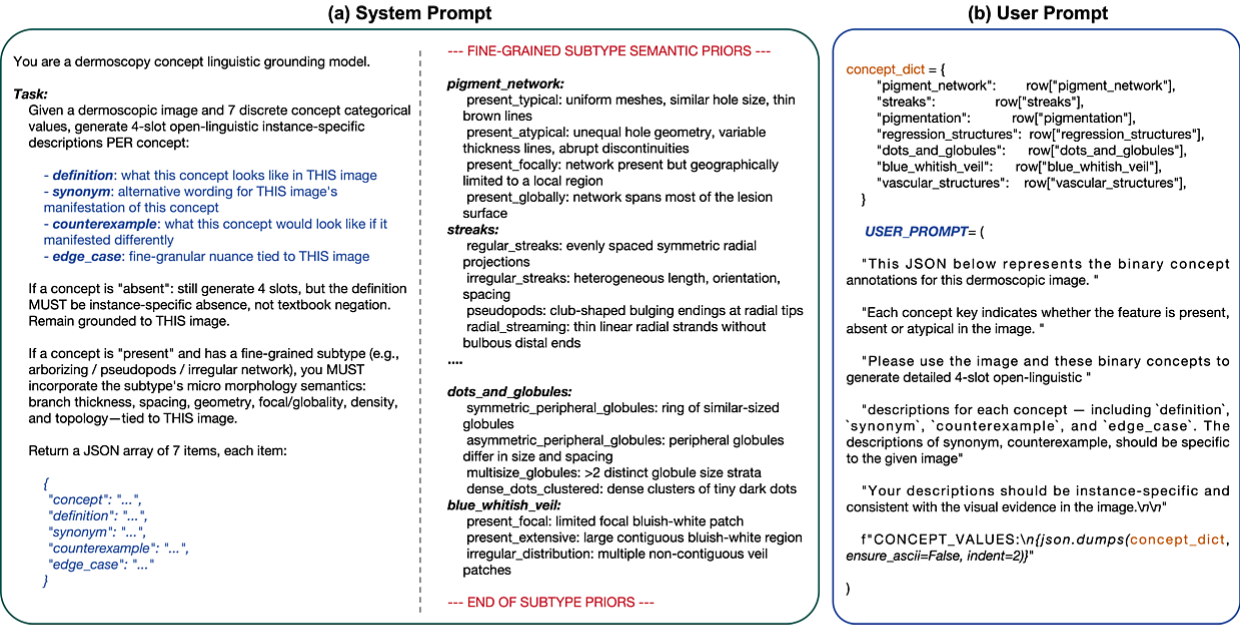}
  \caption*{Figure A2: The prompt engineering architecture for generating Multi-Faceted Semantic Specifications. The System Prompt explicitly incorporates fine-grained subtype semantic priors to guide the LVLM in capturing precise micro-morphology nuances, while the User Prompt dynamically pairs the target image with its categorical concepts.}
  % \vspace{-0.3cm}
  \label{fig:example}
\end{figure*}

Table A2 details the comprehensive hyperparameter configurations for UniCon's two-stage training paradigm. In Stage 1, a relatively large batch size of 128 is employed over 150 epochs to ensure a stable and diverse contrastive learning landscape, which is crucial for robust instance-level semantic alignment. The multi-margin parameters (e.g., $\tau_{close}=0.1$ and $\tau_{spread}=0.9$) are meticulously designed to enforce a strict geometric separation between positive semantic anchors and negative counter-examples in the shared space. 

Furthermore, the three-step joint optimization in Stage 2 is governed by tailored learning rate (LR) schedules to effectively decouple interpretable concept learning from final diagnostic classification. Specifically, while Step 1 optimizes the concept bottleneck at a standard LR of $1 \times 10^{-4}$, Step 2 utilizes a significantly higher LR of $5 \times 10^{-4}$. This allows the diagnostic head to rapidly converge and fit the frozen, purified concept features without overfitting to visual noise. Finally, Step 3 executes end-to-end fine-tuning with a strictly reduced LR of $1 \times 10^{-5}$. This careful learning rate decay prevents catastrophic forgetting of the well-aligned semantic bottleneck, allowing for gentle synergistic adjustments across all modules. Additionally, the balanced loss weights ($\lambda_{align}=\lambda_{cpt}=1.0$) combined with a moderate gating constraint ($\lambda_{gate}=0.5$) ensure that the reliability gate stabilizes dynamically without dominating the core concept prediction.

\section{Prompt and Demonstration of Multi-Faceted Semantic Specifications}

Figure A1 and Figure A2 illustrate the operational pipeline and qualitative outcomes of our Multi-Faceted Semantic Specifications (MSS) generation, a crucial step for establishing language-robust and boundary-sensitive instance-level alignment in the UniCon framework.

\textbf{Figure A1 (Instance-Specific Output Examples):} This figure presents the JSON-formatted linguistic outputs generated by the LVLM. The examples specifically highlight the framework's strict adherence to image-conditioned grounding, even for ``absent'' concepts (such as \textit{streaks} and \textit{regression\_structures} in this sample). Rather than outputting a generic textbook negation, the model details an \textit{instance-specific absence}. For instance, it describes the lesion's margin as ``indistinct but lacks any directional linear elements'' for streaks, and notes that the ``background skin texture remains unaltered by regression phenomena.'' By explicitly characterizing these visual negatives and edge-case nuances directly tied to the current image, UniCon successfully constructs enriched textual specifications. These high-quality semantic anchors effectively bridge the modality gap, laying a trustworthy foundation for the subsequent Unified Concept Prototype Codebook (UCPC) construction.

\textbf{Figure A2 (Prompt Engineering Architecture):} This figure details the comprehensive prompt templates designed to instruct the Large Vision-Language Model (LVLM, specifically Qwen-VL). The architecture is divided into a System Prompt and a User Prompt. To overcome the limitations of sparse textual semantics and ambiguous concept boundaries, the System Prompt mandates the generation of four complementary textual slots: Description ($T^{desc}$), Synonym ($T^{syn}$), Counter-example ($T^{neg}$), and Edge-cases ($T^{edge}$). Crucially, we inject ``Fine-Grained Subtype Semantic Priors'' directly into the instructions. This explicitly guides the LVLM to incorporate micro-morphology semantics (e.g., branch thickness, hole geometry, or melanin distribution) rather than relying on generic vocabulary. The User Prompt then dynamically pairs the specific dermoscopic image with its corresponding concept categorical values to initiate the generation process.

\end{document}